\documentclass{article}

\PassOptionsToPackage{numbers, compress}{natbib}

\usepackage[main, final]{neurips_2026}

\usepackage[utf8]{inputenc} 
\usepackage[T1]{fontenc}    
\usepackage{url}            
\usepackage{booktabs}       
\usepackage{amsfonts}       
\usepackage{nicefrac}       
\usepackage{microtype}      
\usepackage{xcolor}         
\usepackage{graphicx} 
\usepackage{amsmath} 
\usepackage{multirow}
\usepackage{xspace}
\usepackage[inline, shortlabels]{enumitem}
\usepackage{subfig}
\newcommand{\methodName}{\text{FrameScout}\xspace}

\title{\methodName: Scouting Query-Relevant Frames for Long Video Understanding}

\author{%
  Haonan Hu\textsuperscript{1,4}\thanks{Equal contribution. Work by Haonan Hu was done during an internship at Weixin.}\hspace{0.8em}%
  Shuhao Chen\textsuperscript{1,2}\footnotemark[1]\hspace{0.8em}%
  Weisen Jiang\textsuperscript{3}\footnotemark[1]\hspace{0.8em}%
  Lizhao Gao\textsuperscript{5}\hspace{0.8em}%
  James Kwok\textsuperscript{2}\hspace{0.8em}%
  Yu Zhang\textsuperscript{1}\thanks{Corresponding author.} \\
  \textsuperscript{1}Southern University of Science and Technology \\
  \textsuperscript{2}The Hong Kong University of Science and Technology \\
  \textsuperscript{3}South China University of Technology \\
  \textsuperscript{4}Weixin, Tencent Inc. \\
  \textsuperscript{5}Independent Researcher \\
}

\usepackage{microtype}
\usepackage{graphicx}
\usepackage{subfig}
\usepackage{caption}
\usepackage{booktabs} 
\usepackage[pagebackref=false,breaklinks=true,colorlinks,bookmarks=false]{hyperref}
\hypersetup{linkcolor=[rgb]{0.7,0.1,0.1}}
\hypersetup{citecolor=[rgb]{0.4,0.15,0.95}}

\usepackage{algpseudocode}
\usepackage{algorithm}

\usepackage{colortbl}

\usepackage{amsmath}
\usepackage{amssymb}
\usepackage{mathtools}
\usepackage{amsthm}

\usepackage[capitalize,noabbrev]{cleveref}
\usepackage{colortbl}
\usepackage{nicematrix}
\usepackage[inline, shortlabels]{enumitem}
\usepackage{multirow}

\usepackage{amsthm}
\usepackage{amsmath,amsfonts,bm}
\usepackage{nicefrac}
\usepackage{array}
\usepackage{wrapfig}

\usepackage{url}
\usepackage{pbox}

\definecolor{bluex}{rgb}{0.27, 0.42, 0.81}
\definecolor{purplex}{HTML}{9564bf}
\definecolor{red3}{HTML}{C52A20}
\definecolor{red2}{HTML}{B36A6F}
\definecolor{red1}{HTML}{FFb5b5}
\definecolor{purple}{HTML}{B36A6F}
\definecolor{darkyellow}{HTML}{D5BA82}
\definecolor{blue1}{HTML}{508AB2}
\definecolor{blue2}{HTML}{C4E4E3}
\definecolor{green1}{HTML}{A1D0C7}
\definecolor{green2}{HTML}{BFF6BA}
\definecolor{green3}{HTML}{028100}
\definecolor{teal}{HTML}{508AB2}
\definecolor{Gray}{gray}{0.94}
\definecolor{Gray2}{gray}{0.7}
\definecolor{orange3}{HTML}{c28c69}
\definecolor{blue3}{HTML}{3b75af}

\usepackage{pifont}

\usepackage[listings]{tcolorbox}
\tcbuselibrary{skins, listings,theorems,breakable}
\newtcolorbox{mybox}{colback=white!5!white,colframe=black!75!black, left=.05in, right=.05in}

\newtcbtheorem{emptybox}{}%
{colback=green2!5,colframe=blue1,fonttitle=\bfseries,theorem style=plain, left=.0in, right=.0in,bottom=.02in, top=.02in,terminator sign none}{emptybox}
\newtcbtheorem[]{emptyboxx}{A Simple Template For Creating Backward Question}%
{colback=green2!5,colframe=blue1,fonttitle=\bfseries,theorem style=plain, left=.0in, right=.0in,bottom=.02in, top=.02in,terminator sign none}{emptyboxx}
\newtcbtheorem{bth}{Theoreme}
{colback=red!20,colframe=red,theorem style=plain,fonttitle=\bfseries,coltitle=black,terminator sign none}{th}

\definecolor{titlebg}{RGB}{0,0,86}        
\definecolor{framebg}{RGB}{0,0,86}        
\definecolor{bodybg}{RGB}{232,232,238}    

\newtcbtheorem[]{exmp}{Example}{
  enhanced,
  colback=bodybg,
  colframe=framebg,
  coltitle=white,
  colbacktitle=titlebg,
  fonttitle=\bfseries,
  left=.02in,
  right=.02in,
  bottom=.02in,
  top=.02in,
  boxrule=1.2pt,
  arc=2mm
}{exmp}
\newtcbtheorem[]{template}{Template}%
{colback=green2!5,colframe=blue1,fonttitle=\bfseries, left=.02in, right=.02in,bottom=.02in, top=.02in}{template}
\newtcbtheorem[]{prompt}{Prompt}%
{colback=green2!5,colframe=blue1,fonttitle=\bfseries, left=.02in, right=.02in,bottom=.02in, top=.02in}{prompt}
\newtcbtheorem[number within=section]{thm}{Theorem}%
{colback=green!5,colframe=green!35!black,fonttitle=\bfseries}{th}
\newtcbtheorem[number within=section]{corr}{Corollary}%
{colback=green!5,colframe=green!35!black,fonttitle=\bfseries}{th}
\newtcbtheorem{method}{Method}%
{colback=green!5,colframe=green!35!black,fonttitle=\bfseries}{th}

\newtcbtheorem{ebox}{}
{colback=green2!5,colframe=blue1, left=.02in, right=.02in,bottom=.02in,title empty, top=.02in, theorem style=plain apart,frame empty}{ebox}
\theoremstyle{plain}

\theoremstyle{definition}

\theoremstyle{remark}

\newcommand{\ie}{i.e.}
\newcommand{\eg}{e.g.}

\newcommand{\hB}{\mathcal{B}}

\newcommand{\hE}{\mathcal{E}}

\newcommand{\hM}{\mathcal{M}}

\newcommand{\hO}{\mathcal{O}}
\newcommand{\hP}{\mathcal{P}}

\newcommand{\hS}{\mathcal{S}}

\newcommand{\vh}{{\bf h}}

\newcommand{\vC}{{\bf C}}

\newcommand{\vF}{{\bf F}}

\newcommand{\vM}{{\bf M}}

\newcommand{\vtheta}{{\boldsymbol \theta}}

\newif\ifrevhl \revhlfalse
\definecolor{revcolor}{HTML}{028100}
\ifrevhl
  \newcommand{\rev}[1]{{\color{revcolor}#1}}
  \newenvironment{revblock}{\color{revcolor}}{}
\else
  \newcommand{\rev}[1]{#1}
  \newenvironment{revblock}{}{}
\fi

\begin{document}

\maketitle

\begin{abstract}
    Long video understanding requires VideoLLMs to answer queries about videos spanning tens of minutes to hours, but such long videos produce far more visual tokens than current VideoLLMs can process. To address this issue, keyframe selectors are proposed to select the most query-relevant frames by computing the similarity between frame and query embeddings.
    Existing selectors either encode frames independently without temporal context or process chunks in isolation with ranking objectives that are difficult to optimize. To alleviate those limitations, we propose \textbf{\methodName} with two core designs: (i) a \emph{streaming selector architecture} that leverages a sliding-window KV cache and successor aggregation to produce temporally-contextualized frame embeddings; (ii) a \emph{frame-query contrastive objective} that aligns frame and query embeddings in a shared space, directly training the selector to distinguish relevant frames from irrelevant ones.
    Extensive experiments on three long-video benchmarks and six base VideoLLMs demonstrate that \methodName consistently achieves state-of-the-art performance.
    \rev{Code is available at \url{https://github.com/Hubert-1102/FrameScout}.}
\end{abstract}

\section{Introduction}
\label{sec:intro}

Long video understanding requires Video Large Language Models (VideoLLMs)~\citep{chen2024internvl,bai2025qwen3,qwen2.5-VL,zhang2024llava,openai2025gpt5} to answer queries about videos spanning tens of minutes to hours, a task with broad applications from question answering~\citep{xiao2021nextqa,mangalam2023egoschema,videomme} to temporal reasoning~\citep{liu2024tempcompass,cai2024temporalbench,li2024mvbench}.
However, this task poses a fundamental challenge due to the large number of visual tokens, which grows linearly with video length and incurs quadratic attention cost.
For example, a 20-minute video sampled at one frame per second \rev{(256 visual tokens per frame under the Qwen2.5-VL encoder at $448\times448$ resolution)} yields ${\sim}307$K tokens, \rev{exceeding the context window of many VideoLLMs (\eg, 32K tokens for Qwen2.5-VL-7B~\citep{qwen2.5-VL}). Even for recent models with substantially longer contexts (\eg, up to 1M tokens for Qwen3-VL~\citep{bai2025qwen3}), an hour-scale video at 1\,fps saturates the window, and densely prefilling such long contexts incurs prohibitive computation, memory, and latency.}

Uniform downsampling, which reduces the number of frames to fit within the context window, is a naive strategy but inevitably discards query-relevant frames.
This motivates the \textbf{keyframe selection} problem: learning a selector that identifies the most relevant frames for each query.
An effective keyframe selector should satisfy two core requirements: (i)~building \emph{temporally-contextualized frame embeddings} that capture full video context, since determining whether a frame is relevant to a query often requires understanding the temporal context surrounding it; and (ii)~learning a \emph{frame-query alignment} mechanism that accurately selects query-relevant frames from these embeddings.

Existing keyframe selection methods~\citep{tang2025adaptive,yu2024frame,yao2025k,yu2024sevila,wang2024videoagent,fan2024videoagent,zhu2025focus,zhang2025flexselect} fall short of satisfying these two requirements.
For example, FOCUS~\citep{zhu2025focus} adopts a \textit{frozen} vision-language encoder (\eg, BLIP~\citep{li2022blip}) to score each frame \emph{independently} against the query in CLIP space~\citep{radford2021learning}, so the resulting scores lack temporal context across frames, and the frozen encoder provides only surface-level frame-query alignment that is insufficient for long-video understanding.
Recently,
FlexSelect~\citep{zhang2025flexselect} improved upon this by partitioning the video into non-overlapping chunks and training a selector with a Spearman rank-correlation loss~\citep{spearman}.
However, chunks are still processed in isolation, leaving frame representations unaware of cross-chunk temporal context.
The rank-correlation objective supervises full relative ordering among all frames rather than directly aligning frame and query embeddings in a shared space, making it both difficult to optimize and misaligned with the keyframe selection task.

In this paper, we propose \textbf{\methodName}, a keyframe selection framework that \emph{scouts}  query-relevant frames from long videos, addressing the above limitations with two core designs:
(i) a \textbf{streaming selector architecture} that builds temporally-contextualized frame embeddings across the entire video. It combines a sliding-window KV cache that propagates past context across chunk boundaries with a successor aggregation mechanism that incorporates future context, requiring only \emph{constant} KV-cache memory and \emph{linear} computation.
(ii) a \textbf{frame-query contrastive objective} that aligns frame and query embeddings in a shared space via contrastive learning, which pushes relevant frame embeddings closer to the query embedding and pulls irrelevant ones away, directly training the selector to distinguish relevant frames from irrelevant ones. 
Across three long-video benchmarks (i.e., VideoMME~\citep{videomme}, MLVU~\citep{zhou2025mlvu}, LongVB~\citep{wu2024longvideobench}) and six base VideoLLMs (both closed- and open-source), \methodName consistently establishes state-of-the-art performance, \eg, $+6.5\%$ on VideoMME-Long with Qwen3-VL-8B~\citep{bai2025qwen3}, while achieving a $6\times$ speedup in multi-query scenarios.

The main contributions of this paper are as follows:
\begin{itemize}[leftmargin=10pt, itemsep=0pt,topsep=0pt]
    \item We propose {\methodName}, a keyframe selection framework built on a streaming selector architecture that produces temporally-contextualized frame representations via a sliding-window KV cache and successor aggregation, under constant KV-cache memory and linear computation in video length.
    \item We design a frame-query contrastive objective that aligns frame and query embeddings in a shared space, directly training the selector to distinguish query-relevant frames from irrelevant ones.
    \item Extensive experiments on three long-video understanding benchmarks and six base VideoLLMs demonstrate that \methodName consistently outperforms the current state-of-the-art selectors FOCUS~\citep{zhu2025focus} and FlexSelect~\citep{zhang2025flexselect}
    with a $6\times$ speedup in multi-query scenarios.
\end{itemize}


\section{Related Works}
\label{sec:related_works}

\textbf{Keyframe Selection for Long Video Understanding.}
Early Video Large Language Models (VideoLLMs)~\citep{zhang2024llava,chen2024internvl,bai2025qwen3,qwen2.5-VL} extend
image-based multimodal LLMs by densely encoding a small number of sampled frames,
yielding strong performance on short video benchmarks but scaling poorly with
video duration.
Since long videos produce far more visual tokens than current VideoLLMs can
process, selecting a compact set of query-relevant keyframes has emerged as a
practical strategy. 
Existing keyframe selection methods can be grouped into three families.
\begin{enumerate*}[(i), series = tobecont, itemjoin = \quad]
    \item \textit{Heuristic methods}, such as uniform sampling and visual-similarity
    clustering, are training-free but often miss important moments and cannot
    condition on the query~\citep{tang2025adaptive}.
    \item \textit{Similarity-based methods} leverage frozen pre-trained encoders to score
    frames by cross-modal matching. For example, FOCUS~\citep{zhu2025focus} uses CLIP~\citep{radford2021learning}/BLIP~\citep{li2022blip} for
    text-image similarity, and Adaptive Greedy~\citep{tang2025adaptive} combines
    SigLIP~\citep{zhai2023sigmoid} and DINOv2~\citep{oquab2024dinov2} under a submodular objective to balance relevance and coverage.
    \rev{BOLT~\citep{liu2025bolt} applies inverse transform sampling to CLIP query-frame similarities, and F2C~\citep{sun2025f2c} extends frame selection to key clips with adaptive resolution under a fixed token budget.}
    Although efficient, these methods score each frame \emph{independently} in a
    \emph{frozen} embedding space: frame representations carry no temporal context,
    and the pretrained CLIP-style alignment captures only surface-level concept
    overlap, insufficient for long-video QA.
    \item \textit{Learning-based methods}~\citep{yu2024frame,zhang2025flexselect,hu2025mllm,yao2025k} like FlexSelect~\citep{zhang2025flexselect} train a 
    lightweight selector to rank frames according to their relevance to the query.
    However, they partition the video into non-overlapping chunks processed in
    isolation, leaving frame representations blind to cross-chunk temporal context.
    Their ranking objectives (\eg, the Spearman rank-correlation loss used by
    FlexSelect~\citep{zhang2025flexselect}) also entangle frame encoding with the
    query, requiring the entire selector to be re-run for every new query.
\end{enumerate*}

The proposed \methodName differs from the above methods in three respects:
(i)~frame representations capture cross-chunk temporal context rather than encoding frames or chunks in isolation;
(ii)~the training objective targets relevant-vs-irrelevant separation rather than full ranking;
(iii)~frame embeddings are query-agnostic rather than entangled with query encoding.

\textbf{Efficient Long-Context Modeling.}
Processing long sequences under bounded resources has been extensively studied.
Sliding-window attention~\citep{beltagy2020longformer,jiang2023mistral} restricts
each token to attend only to a fixed-size local neighborhood.
StreamingLLM~\citep{xiao2024efficient} further identifies the \emph{attention sink}
phenomenon and retains a small set of initial tokens alongside the sliding window
to stabilize the attention distribution, enabling unbounded generation under a
bounded KV cache; subsequent work~\citep{han2024lminfinite,cai2024pyramidkv}
extends this idea with hierarchical caches and dynamic eviction.
These techniques have been developed primarily for text-only LLMs;
\methodName is the first to adopt sliding-window KV caching for keyframe selection, enabling constant-memory streaming over arbitrarily
long videos.

However, sliding-window caching is inherently \emph{causal}: each position only
sees past context, but keyframe selection requires temporally-contextualized
frame representations that also capture future context.
Existing methods such as bidirectional
encoders~\citep{devlin2019bert,schuster1997bidirectional} provide full-context
awareness but require KV-cache memory that grows linearly with sequence length, making them impractical for long videos.
To incorporate future context without breaking the streaming property, \methodName
draws on the idea of successor representations in reinforcement
learning~\citep{dayan1993successor,barreto2017successor}, which express a
state's value as an exponentially-discounted aggregation of future rewards.
Unlike bidirectional encoders that require memory linear in sequence length,
this \emph{successor aggregation} accumulates an exponentially-decayed summary of
all subsequent frame embeddings via a single backward pass, achieving
bi-directional contextualization under constant KV-cache memory and linear time.

\section{Methodology}
\label{sec:method}

\begin{figure}[!t]
    \centering
    \includegraphics[width=1.0\linewidth]{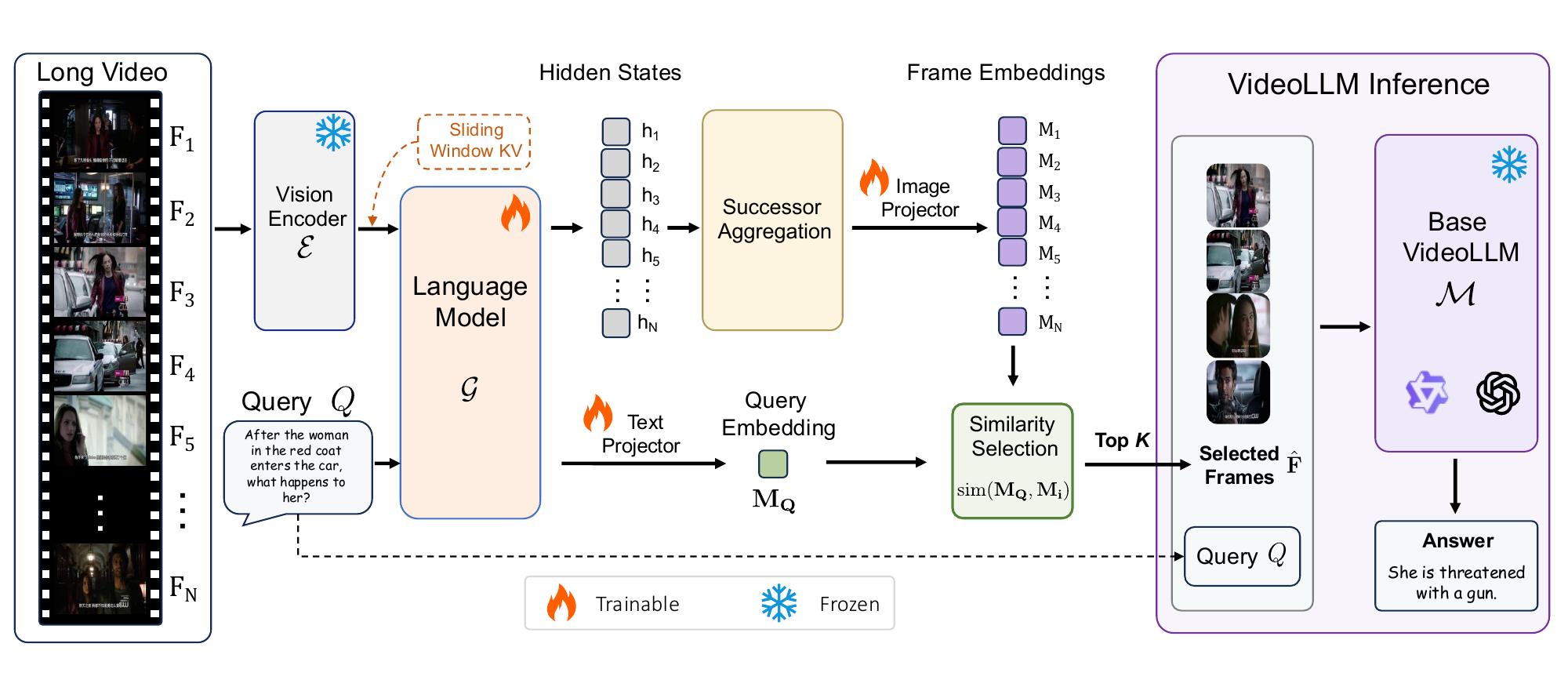}
    \vskip -.1in
    \caption{Illustration of \methodName.}
    \label{fig:framework}
    \vskip -.1in
\end{figure}

Current VideoLLMs are constrained by fixed context windows, making it essential to select a compact set of query-relevant keyframes from long videos. This requires both temporally-contextualized frame representations that capture the full video context, and a learned alignment between frame and query semantics. \methodName achieves this with two core components (Figure~\ref{fig:framework}): a \textbf{streaming selector architecture} (Section~\ref{sec:memory}) that builds temporally-contextualized frame representations, and a \textbf{frame-query contrastive objective} (Section~\ref{sec:training}) that learns accurate frame-query alignment for keyframe selection.
We begin by formalizing the problem in Section~\ref{sec:problem}.

\subsection{Problem Formulation}
\label{sec:problem}

Given a long video $\vF = (F_1, \ldots, F_N)$ of $N$ frames and a text query $Q$, our goal is to learn a lightweight selector $\mathcal{S}_\vtheta(\vF, Q) \rightarrow \hat{\vF}$ that
 chooses a $K$-frame subsequence $\hat{\vF} = (F_{i_1}, \ldots, F_{i_K})$ from $\vF$ maximizing the answer quality of a downstream base VideoLLM $\hM$:
\begin{equation}
    \hat{\vF} = \arg\max_{|\hat{\vF}|=K} \; \hP\big(\hM(\hat{\vF}, Q), A\big),
\end{equation}
where $A$ is the ground-truth answer and $\hP$ is the answer quality metric.

The selector must be much smaller than the base VideoLLM $\hM$; otherwise, the cost of selection would exceed that of processing the full video directly with $\hM$.
Designing such a selector poses two challenges: \textbf{(i)}~a smaller model cannot process all frames at once, so how to build \textbf{ temporally-contextualized frame representations} that capture full video context (Section~\ref{sec:memory}), and \textbf{(ii)}~how to train the selector so that \textbf{query-relevant frames} are scored above irrelevant ones (Section~\ref{sec:training}).
 
\subsection{Streaming Selector Architecture}
\label{sec:memory}

Keyframe selection requires temporally-contextualized frame representations that capture the full video context. 
Although an autoregressive VideoLLM can contextualize every frame with all preceding frames via causal attention, 
processing all frames at once far exceeds its context window for a long video.
Existing methods either encode each frame independently with vision-language encoders~\citep{zhu2025focus, tang2025adaptive}, or partition the video into chunks and encode frames jointly within each chunk~\citep{yu2024frame, zhang2025flexselect}; in both cases, frames or chunks are processed in isolation, so the resulting representations lack cross-boundary temporal context and future context.

We address this limitation by designing the selector $\mathcal{S}_\vtheta$ as a streaming architecture, using a sliding-window KV cache to incorporate past context and successor aggregation to incorporate future context. 
Specifically, $\mathcal{S}_\vtheta$ is a lightweight VideoLLM comprising a vision encoder $\mathcal{E}$, a language model $\mathcal{G}$, and a two-layer MLP projector $\phi$.
Given the input video $\vF = (F_1, \ldots, F_N)$, 
$\hE$ first encodes each frame into visual tokens: $\mathbf{v}_i = \mathcal{E}(F_i) \in \mathbb{R}^{L \times d}$, where $L$ is the number of tokens per frame and $d$ is the hidden dimension.
Like existing methods, the selector partitions the video into non-overlapping chunks of $B$ frames to fit within the context window. Let $\mathcal{B}_t = (F_{(t-1)B+1}, \ldots, F_{tB})$ denote the $t$-th chunk ($t = 1, \ldots, \lceil N/B \rceil$) and $\mathbf{V}_t = (\mathbf{v}_{(t-1)B+1}, \ldots, \mathbf{v}_{tB})$ the corresponding visual tokens.
Unlike existing methods, however, our selector can maintain cross-boundary temporal context by using a sliding-window KV cache and successor aggregation.

\begin{figure}[!t]
    \centering
    \includegraphics[width=1.0\linewidth]{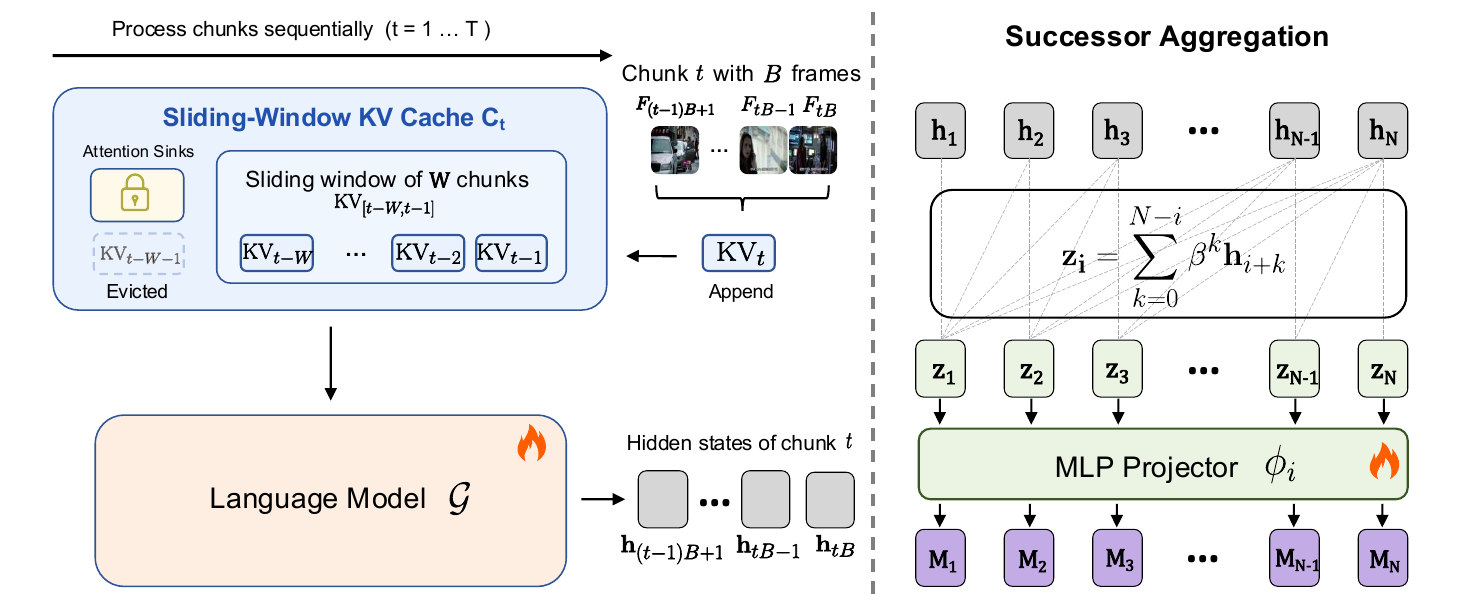}
    \caption{Illustration of the sliding-window KV Cache and successor aggregation mechanisms.}
    \label{fig:sliding_window}
\end{figure}

\paragraph{History context via sliding-window KV cache.}
As illustrated in Figure~\ref{fig:sliding_window}, the sliding-window cache consists of two parts: (i)~a fixed system prompt $\mathbf{s}$, whose KV cache $\text{KV}(\mathbf{s})$ is computed once and always retained as \emph{attention sinks}~\citep{xiao2024efficient} to stabilize the attention distribution; and (ii)~a sliding window of size $W$ storing the KV cache of the $W$ most recently processed chunks, denoted $\text{KV}_{[t-W,\, t-1]}$.
The full cache when processing chunk $t$ is $\vC_t = (\text{KV}(\mathbf{s}),\, \text{KV}_{[t-W,\, t-1]})$.
The backbone $\mathcal{G}$ takes $\mathbf{V}_t$ and $\vC_t$, producing a hidden state $\vh_i \in \mathbb{R}^{d}$ for each frame $F_i$ in the chunk:
\begin{equation}
    (\vh_{(t-1)B+1}, \ldots, \vh_{tB}) =
    \mathcal{G}\big(\vC_t,\, \mathbf{V}_t\big).
\end{equation}
After processing chunk $\mathcal{B}_t$, the window slides forward: $\mathbf{V}_t$'s KV entries (i.e., $\text{KV}_{t}$) are appended and the oldest chunk's entries $\text{KV}_{t-W}$ are evicted, while $\text{KV}(\mathbf{s})$ is always kept. The remaining entries are re-indexed via rotary position embedding (RoPE) to preserve correct positional information. The total cache size stays constant at $|\mathbf{s}| + W B L$ tokens regardless of video length $N$.

Through this cache chain, each $\vh_i$ is contextualized by more than the $W B$ frames in the current window: each earlier chunk's hidden states were computed while attending to their own preceding context, and this information propagates forward through successive KV states. By induction, $\vh_i$ is informed by all preceding frames $F_1, \ldots, F_i$.

\paragraph{Future context via successor aggregation.}
The KV-cache chain contains rich past context, but $\vh_i$ cannot see future context. To incorporate future information, we propose successor aggregation to encode future context into frame representations with exponential temporal discounting, so that nearby frames contribute more than distant ones:
\begin{equation}
    \vM_i
    = \phi \left(\sum_{k=0}^{N-i}\beta^{k}\,\vh_{i+k}\right)
    \in \mathbb{R}^{d},
\end{equation}
where $\beta \in [0,1]$ is a discount factor. Letting $\mathbf{z}_i = \sum_{k=0}^{N-i}\beta^{k}\,\vh_{i+k}$, we have the recurrence $\mathbf{z}_i = \vh_i + \beta\,\mathbf{z}_{i+1}$ with $\mathbf{z}_N = \vh_N$, so the entire sequence $\mathbf{z}_N, \ldots, \mathbf{z}_1$ can be computed by a single backward pass. 
This recurrence shares a similar form with the successor representation~\citep{dayan1993successor, kulkarni2016deep} in reinforcement learning, but is applied here to aggregate future context rather than to decompose a value function.
The resulting frame embedding $\vM_i = \phi(\mathbf{z}_i)$ captures the full temporal context: past frames through the KV-cache chain, and future frames through the successor aggregation.
After processing all $N$ frames, we obtain their embeddings $(\vM_1, \ldots, \vM_N)$.

\textbf{Complexity analysis.}
Let $L_s = |\mathbf{s}|$ denote the system prompt length and $T = \lceil N/B \rceil$ the total number of chunks. In a standard full-context VideoLLM, processing all $N$ frames simultaneously yields a sequence of $N L$ tokens, incurring $\hO(N^2 L^2 d)$ computation and $\hO(NLd)$ memory for the KV cache. In contrast, \methodName processes $N$ frames via streaming, with the attention context at each step bounded by $L_s + W B L$ tokens:
\begin{equation}
    \text{Computation} = \hO\big(T (L_s + WBL)^2 d\big), \quad \text{Memory} = \hO\big((L_s + WBL)\, d\big).
\end{equation}
Since $W$ and $B$ are fixed constants with $WB \ll N$, the computational cost is \textit{linear} in video length and the KV-cache memory is \textit{constant}. The successor aggregation recurrence adds only $O(Nd)$ computation and $O(d)$ memory, which is dominated by the attention cost above. Together, these properties enable \methodName to process videos of arbitrary duration.

\subsection{Frame-Query Contrastive Learning}
\label{sec:training}

The streaming architecture produces query-agnostic frame embeddings $(\vM_1, \ldots, \vM_N)$. To enable query-relevant frame selection, we propose a frame-query contrastive objective that aligns frame and query embeddings in a shared space, so that their similarity reflects frame-query relevance.

\subsubsection{Training}
\label{sec:training_detail}

Existing video QA datasets only provide video-level question-answer pairs, not frame-level relevance annotations. 
Following~\citep{zhang2025flexselect}, we generate pseudo-labels by feeding a sampled subset of $N'$ frames ($N'{=}100$) into a teacher VideoLLM (\eg, the base VideoLLM) and extracting the attention weights from query tokens to visual tokens at a reference layer. 
These weights are averaged across heads and aggregated per frame to obtain a query-frame relevance score. Note that $N'$ is chosen to fit within the teacher's context window, enabling a single forward pass. Based on these scores, we select the top-$k$ frames as the positive set $\mathcal{T}_+$ and treat all remaining frames as the negative set $\mathcal{T}_-$.

The query $Q$ is encoded through the same backbone $\mathcal{G}$ and projector $\phi$ used for frame embeddings, yielding a query embedding $\vM_Q = \phi(\mathcal{G}(Q)) \in \mathbb{R}^{d}$.
Let $\operatorname{sim}(\vM_Q, \vM_i)$ denote the cosine similarity between $\vM_Q$ and frame embedding $\vM_i$.
We train the selector with a frame-query contrastive loss:
\begin{equation}
    \mathcal{L} = -\frac{1}{|\mathcal{T}_+|}
      \sum_{i \in \mathcal{T}_+} \log
      \frac{\exp(\operatorname{sim}(\vM_Q, \vM_i) / \tau)}
           {\exp(\operatorname{sim}(\vM_Q, \vM_i) / \tau) + \displaystyle\sum_{j \in \mathcal{T}_-}
            \exp(\operatorname{sim}(\vM_Q, \vM_j) / \tau)},
    \label{eq:infonce}
\end{equation}
where $\tau$ is a temperature hyperparameter.
We choose the contrastive loss over the Spearman correlation loss used in FlexSelect~\citep{zhang2025flexselect} because keyframe selection only requires relevant frames to rank \emph{above} irrelevant ones, not the relative ordering among frames.
The contrastive loss directly optimizes this relevant-vs-irrelevant separation, matching the criterion used at inference.

Algorithm~\ref{alg:train} in Appendix \ref{apd:algorithms} summarizes the training procedure.
During training, we process each chunk independently with only the
sink KV cache $\text{KV}(\mathbf{s})$, because this enables
chunk-level parallelism and bounds per-sample activation memory to a
single chunk, making training efficient and scalable.
Cross-chunk temporal context is still captured by the successor
aggregation, which operates identically to inference and keeps the
learned representations consistent.
The streaming sliding-window KV cache is activated at inference to further enrich frame representations with history context.
\rev{Section~\ref{sec:train_context} verifies that, under the same wall-clock budget, this strategy converges faster and is more accurate than training with the streaming KV context.}

\subsubsection{Inference}
\label{sec:inference}

Since frame embeddings are query-agnostic, they are computed once per video and reused across all queries.
As shown in Algorithm~\ref{alg:infer} of Appendix \ref{apd:algorithms},
the inference pipeline is separated into two steps:

\textbf{Step~1: Video encoding (once per video).}
The selector streams through the entire video as described in Section~\ref{sec:memory}, producing frame embeddings $(\vM_1, \ldots, \vM_N)$.

\textbf{Step~2: Frame selection (per query).}
For a query $Q$, we compute its embedding $\vM_Q = \phi(\mathcal{G}(Q))$ and select the top-$K$ frames:
\begin{equation}
    \hat{\mathcal{I}} = \operatorname{Top\text{-}K}
      \big(\{\operatorname{sim}(\vM_Q,\, \vM_i)\}_{i=1}^{N}\big),
    \quad
    \hat{\vF} = (F_i)_{i \in \hat{\mathcal{I}}},
    \label{eq:topk}
\end{equation}
where $\operatorname{Top\text{-}K}$ returns the indices of the $K$ highest-scoring frames. 
The selected keyframes $\hat{\vF}$ are then fed into the base VideoLLM $\hM$ with $Q$ to produce the answer $A = \hM(\hat{\vF},\, Q)$.
Step~2 involves only one text forward pass and $N$ dot products, so the per-query cost is negligible (${\sim}$0.03\,s; Section~\ref{sec:expt:main}). In contrast, existing methods~\citep{zhu2025focus, yu2024frame, yao2025k} must re-encode the full video for every new query. 
The selector is also fully decoupled from $\hM$, so $\hM$ can be replaced by any VideoLLM
 (e.g., open-source  Qwen3-VL~\citep{bai2025qwen3} or 
closed-source GPT-5.4~\citep{openai2025gpt5}).

\section{Experiments}
\label{sec:experiments}

\subsection{Experimental Settings}
\label{subsec:exp_settings}

\textbf{Benchmarks.}
Following~\citep{zhang2025flexselect}, we evaluate on three long-video understanding benchmarks:
\begin{enumerate*}[(i), series = tobecont, itemjoin = \quad]
    \item \textbf{VideoMME}~\citep{videomme} provides comprehensive
      evaluation across short, medium, and long videos; we report the
      Long and Overall accuracy without subtitles.
    \item \textbf{MLVU}~\citep{zhou2025mlvu} is a multi-task benchmark
      for long-form video understanding; we report the M-Avg accuracy.
    \item \textbf{LongVB}~\citep{wu2024longvideobench} targets
      retrieval and reasoning in long-context videos; we report the
      validation set accuracy.
\end{enumerate*}

\textbf{Base VideoLLMs.}
The evaluation is conducted on a diverse set of closed- and open-source
base VideoLLMs $\hM$.
Closed-source models include GPT-5.4-mini and
GPT-5.4~\citep{openai2025gpt5}.
Open-source models include
Qwen2.5-VL-7B~\citep{qwen2.5-VL},
LLaVA-Video-7B~\citep{zhang2024llava},
InternVL2.5-8B~\citep{chen2024internvl},
and Qwen3-VL-8B~\citep{bai2025qwen3}.
For the lightweight selector $\mathcal{S}_\vtheta$, we adopt InternVL2.5-1B~\citep{chen2024internvl}
across all base VideoLLMs.

\textbf{Implementation Details.}
We use InternVL2.5-8B as the teacher model to generate attention-based
pseudo-labels on the LLaVA-Video dataset~\citep{zhang2024llava}
(${\sim}$80k video--text pairs).
The temperature $\tau$ is set to 0.05.
The chunk size is $B{=}12$ frames for an accuracy-efficiency trade-off (as shown in Appendix~\ref{subsubsec:sensitivity}), the system-prompt
length is $|\mathbf{s}|{=}256$ tokens, and the discount factor is
$\beta{=}0.3$ (see Appendix~\ref{subsubsec:sensitivity} for a sensitivity analysis).

\begin{table}[!htbp]
    \centering
    \caption{Accuracy (\%) on long-video understanding benchmarks. Relative improvements over uniform sampling are shown in parentheses.}
    \label{tab:main_results}
    \renewcommand{\arraystretch}{1.1} 
    \resizebox{\linewidth}{!}{
    \begin{NiceTabular}{clccccc}
    \CodeBefore
    \rectanglecolor{Gray}{6-3}{6-7}
    \rectanglecolor{Gray}{10-3}{10-7}
    \rectanglecolor{Gray}{14-3}{14-7}
    \rectanglecolor{Gray}{18-3}{18-7}
    \rectanglecolor{Gray}{22-3}{22-7}
    \rectanglecolor{Gray}{26-3}{26-7}
\Body
    \toprule
    \multirow{2}{*}{} & \multirow{2}{*}{\textbf{VideoLLM}} & \multirow{2}{*}{\textbf{Method}} & \multicolumn{2}{c}{\textbf{VideoMME}} & \textbf{MLVU} & \textbf{LongVB} \\
    \cmidrule(lr){4-5} \cmidrule(lr){6-6} \cmidrule(lr){7-7}
    & & & Long & Overall & M-Avg & Val \\
    \midrule
    \multirow{8}{*}{\rotatebox[origin=c]{90}{\textit{Closed-Source}}}
    & \multirow{4}{*}{GPT-5.4-mini~\citep{openai2025gpt5}} & Uniform & 46.9 & 56.1 & 57.8 & 52.9 \\
    & & FOCUS~\citep{zhu2025focus} & 42.1 \textcolor{gray}{\scriptsize{(-4.8)}} & 53.2 \textcolor{gray}{\scriptsize{(-2.9)}} & 62.3 \textcolor{gray}{\scriptsize{(+4.5)}} & 54.2 \textcolor{gray}{\scriptsize{(+1.3)}} \\
    & & FlexSelect~\citep{zhang2025flexselect} & 50.4 \textcolor{gray}{\scriptsize{(+3.5)}} & 59.3 \textcolor{gray}{\scriptsize{(+3.2)}} & 61.2 \textcolor{gray}{\scriptsize{(+3.4)}} & 58.8 \textcolor{gray}{\scriptsize{(+5.9)}} \\
    & & {\methodName} & \textbf{52.3} \textcolor{blue}{\scriptsize{(+5.4)}} & \textbf{60.8} \textcolor{blue}{\scriptsize{(+4.7)}} & \textbf{62.7} \textcolor{blue}{\scriptsize{(+4.9)}} & \textbf{58.9} \textcolor{blue}{\scriptsize{(+6.0)}} \\
    \cmidrule(l){2-7}
    & \multirow{4}{*}{GPT-5.4~\citep{openai2025gpt5}} & Uniform & 60.9 & 66.9 & 68.5 & 63.6 \\
    & & FOCUS~\citep{zhu2025focus} & 54.2 \textcolor{gray}{\scriptsize{(-6.7)}} & 64.9 \textcolor{gray}{\scriptsize{(-2.0)}} & 67.3 \textcolor{gray}{\scriptsize{(-1.2)}} & 63.2 \textcolor{gray}{\scriptsize{(-0.4)}} \\
    & & FlexSelect~\citep{zhang2025flexselect} & {64.7} \textcolor{gray}{\scriptsize{(+3.8)}} & 70.6 \textcolor{gray}{\scriptsize{(+3.7)}} & 64.7 \textcolor{gray}{\scriptsize{(-3.8)}} & 68.1 \textcolor{gray}{\scriptsize{(+4.5)}} \\
    & & {\methodName} & \textbf{65.4} \textcolor{blue}{\scriptsize{(+4.5)}} & \textbf{72.5} \textcolor{blue}{\scriptsize{(+5.6)}} & \textbf{72.0} \textcolor{blue}{\scriptsize{(+3.5)}} & \textbf{68.4} \textcolor{blue}{\scriptsize{(+4.8)}} \\
    \midrule
    \multirow{16}{*}{\rotatebox[origin=c]{90}{\textit{Open-Source}}}
    & \multirow{4}{*}{Qwen2.5-VL-7B~\citep{qwen2.5-VL}} & Uniform & 51.4 & 62.6 & 64.9 & 59.0 \\
    & & FOCUS~\citep{zhu2025focus} & 54.3 \textcolor{gray}{\scriptsize{(+2.9)}} & 63.4 \textcolor{gray}{\scriptsize{(+0.8)}} & 69.3 \textcolor{gray}{\scriptsize{(+4.4)}} & 61.1 \textcolor{gray}{\scriptsize{(+2.1)}} \\
    & & FlexSelect~\citep{zhang2025flexselect} & 55.4 \textcolor{gray}{\scriptsize{(+4.0)}} & 64.9 \textcolor{gray}{\scriptsize{(+2.3)}} & 67.2 \textcolor{gray}{\scriptsize{(+2.3)}} & {63.0} \textcolor{gray}{\scriptsize{(+4.0)}} \\
    & & {\methodName} & \textbf{57.3} \textcolor{blue}{\scriptsize{(+5.9)}} & \textbf{65.6} \textcolor{blue}{\scriptsize{(+3.0)}} & \textbf{70.3} \textcolor{blue}{\scriptsize{(+5.4)}} & \textbf{63.3} \textcolor{blue}{\scriptsize{(+4.3)}} \\
    \cmidrule(l){2-7}
    & \multirow{4}{*}{LLaVA-Video-7B~\citep{zhang2024llava}} & Uniform & 53.5 & 63.8 & 71.5 & 60.4 \\
    & & FOCUS~\citep{zhu2025focus} & 46.9 \textcolor{gray}{\scriptsize{(-6.6)}} & 59.2 \textcolor{gray}{\scriptsize{(-4.6)}} & 69.3 \textcolor{gray}{\scriptsize{(-2.2)}} & 63.2 \textcolor{gray}{\scriptsize{(+2.8)}} \\
    & & FlexSelect~\citep{zhang2025flexselect} & 51.8 \textcolor{gray}{\scriptsize{(-1.7)}} & 62.6 \textcolor{gray}{\scriptsize{(-1.2)}} & 70.2 \textcolor{gray}{\scriptsize{(-1.3)}} & 62.5 \textcolor{gray}{\scriptsize{(+2.1)}} \\
    & & {\methodName} & \textbf{56.1} \textcolor{blue}{\scriptsize{(+2.6)}} & \textbf{65.7} \textcolor{blue}{\scriptsize{(+1.9)}} & \textbf{72.3} \textcolor{blue}{\scriptsize{(+0.8)}} & \textbf{63.4} \textcolor{blue}{\scriptsize{(+3.0)}} \\
    \cmidrule(l){2-7}
    & \multirow{4}{*}{InternVL2.5-8B~\citep{chen2024internvl}} & Uniform & 51.4 & 62.7 & 68.9 & 61.8 \\
    & & FOCUS~\citep{zhu2025focus} & 54.2 \textcolor{gray}{\scriptsize{(+2.8)}} & 61.9 \textcolor{gray}{\scriptsize{(-0.8)}} & 69.5 \textcolor{gray}{\scriptsize{(+0.6)}} & 63.9 \textcolor{gray}{\scriptsize{(+2.1)}} \\
    & & FlexSelect~\citep{zhang2025flexselect} & 55.0 \textcolor{gray}{\scriptsize{(+3.6)}} & 63.3 \textcolor{gray}{\scriptsize{(+0.6)}} & 69.6 \textcolor{gray}{\scriptsize{(+0.7)}} & 65.0 \textcolor{gray}{\scriptsize{(+3.2)}} \\
    & & {\methodName} & \textbf{55.4} \textcolor{blue}{\scriptsize{(+4.0)}} & \textbf{64.3} \textcolor{blue}{\scriptsize{(+1.6)}} & \textbf{70.2} \textcolor{blue}{\scriptsize{(+1.3)}} & \textbf{65.7} \textcolor{blue}{\scriptsize{(+3.9)}} \\
    \cmidrule(l){2-7}
    & \multirow{4}{*}{Qwen3-VL-8B~\citep{bai2025qwen3}} & Uniform & 54.7 & 65.1 & 67.9 & 62.9 \\
    & & FOCUS~\citep{zhu2025focus} & 57.5 \textcolor{gray}{\scriptsize{(+2.8)}} & 64.8 \textcolor{gray}{\scriptsize{(-0.3)}} & 70.7 \textcolor{gray}{\scriptsize{(+2.8)}} & 64.4 \textcolor{gray}{\scriptsize{(+1.5)}} \\
    & & FlexSelect~\citep{zhang2025flexselect} & 60.3 \textcolor{gray}{\scriptsize{(+5.6)}} & 67.7 \textcolor{gray}{\scriptsize{(+2.6)}} & 71.6 \textcolor{gray}{\scriptsize{(+3.7)}} & 64.1 \textcolor{gray}{\scriptsize{(+1.2)}} \\
    & & {\methodName} & \textbf{61.2} \textcolor{blue}{\scriptsize{(+6.5)}} & \textbf{69.3} \textcolor{blue}{\scriptsize{(+4.2)}} & \textbf{72.6} \textcolor{blue}{\scriptsize{(+4.7)}} & \textbf{67.8} \textcolor{blue}{\scriptsize{(+4.9)}} \\
    \bottomrule
    \end{NiceTabular}
    }
\end{table}

\textbf{Baselines.}
We compare \methodName{} against three frame-selection methods:
(i)~\textbf{Uniform}: uniformly samples a subset of frames from the video.
(ii)~\textbf{FOCUS}~\citep{zhu2025focus}: a training-free method that
encodes each frame independently and scores frame--query relevance in CLIP
space, without any temporal modeling.
(iii)~\textbf{FlexSelect}~\citep{zhang2025flexselect}: a learned selection
method that partitions the video into independent chunks and scores frames within each chunk in isolation, preventing information flow across
chunk boundaries.
For all settings and baselines, the base VideoLLM receives exactly $K=60$ selected frames
so that comparisons isolate frame-selection quality from input budget
differences.
\rev{Comparisons with the recent methods BOLT~\citep{liu2025bolt}, F2C~\citep{sun2025f2c}, M-LLM~\citep{hu2025mllm}, HICom~\citep{liu2025hicom}, and VTM~\citep{lee2024vtm} are provided in Appendix~\ref{apd:recent_methods}.}

\subsection{Main Results}
\label{sec:expt:main}

As shown in Table~\ref{tab:main_results}, \methodName{} consistently
outperforms all baselines across three benchmarks and six base VideoLLMs
spanning both closed- and open-source families.

\textbf{Consistent improvements over uniform sampling.}
\methodName{} yields consistent gains over the uniform baseline on every
model--benchmark combination.
The improvements are large on Qwen3-VL-8B
($+6.5$ on VideoMME Long, $+4.7$ on MLVU,
$+4.9$ on LongVideoBench) and Qwen2.5-VL-7B
($+5.9$ on VideoMME Long, $+5.4$ on MLVU).
Crucially, \methodName{} also benefits the strongest base model:
when paired with GPT-5.4, it lifts VideoMME Overall by $+5.6$
and MLVU by $+3.5$, confirming that our selector provides
orthogonal improvements to frontier VideoLLMs.

\textbf{Surpassing state-of-the-art.}
\methodName{} consistently outperforms both FOCUS and FlexSelect across
all base models and benchmarks.
On Qwen3-VL-8B, for example, \methodName{} surpasses FlexSelect by
$+1.6$ on VideoMME Overall, $+1.0$ on MLVU, and $+3.7$ on LongVideoBench;
the margin over FOCUS is even larger.
Moreover, both baselines exhibit inconsistent behavior:
FOCUS frequently degrades performance
(\eg, $-6.7$ on VideoMME Long for GPT-5.4;
$-6.6$ for LLaVA-Video-7B),
and FlexSelect still harms several configurations
(\eg, $-3.8$ on MLVU for GPT-5.4;
$-1.7$ on VideoMME Long for LLaVA-Video-7B).
In contrast, \methodName{} delivers positive gains in all settings.
\rev{\methodName{} also outperforms BOLT, F2C, M-LLM, HICom and VTM (Appendix~\ref{apd:recent_methods}).}
We attribute this robustness to two design choices:
(i) our streaming architecture
contextualizes every frame with the full
video via a sliding-window KV cache and successor aggregation, whereas
existing selectors process frames or chunks in isolation and thus lack
cross-boundary temporal context.
(ii) our frame-query contrastive objective directly optimizes the separation between
relevant and irrelevant frames in a shared embedding space,
enabling more accurate selection than the CLIP-space heuristic\rev{s} of
FOCUS\rev{, BOLT, and F2C}, the rank-supervised per-chunk scoring of FlexSelect\rev{, or the pointwise score regression of M-LLM}.

\begin{figure}[!t]
    \centering
    \begin{minipage}[c]{0.58\textwidth}
        \centering
        \includegraphics[width=\linewidth]{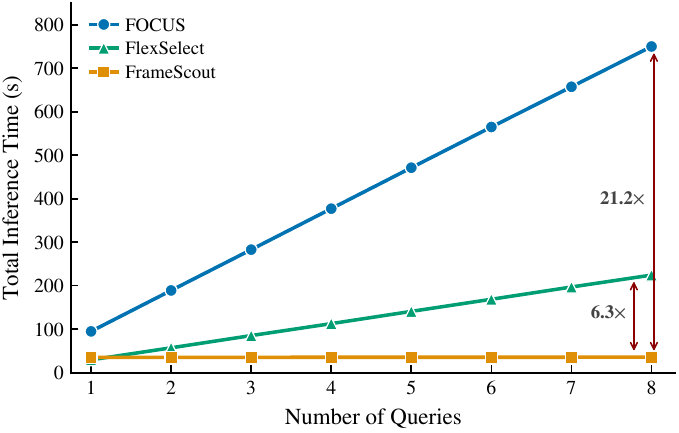}
        \captionof{figure}{Total inference time vs.\ \#Queries.}
        \label{fig:efficiency_comparison}
    \end{minipage}
    \hfill
    \begin{minipage}[c]{0.36\textwidth}
        \centering
        \includegraphics[width=\linewidth]{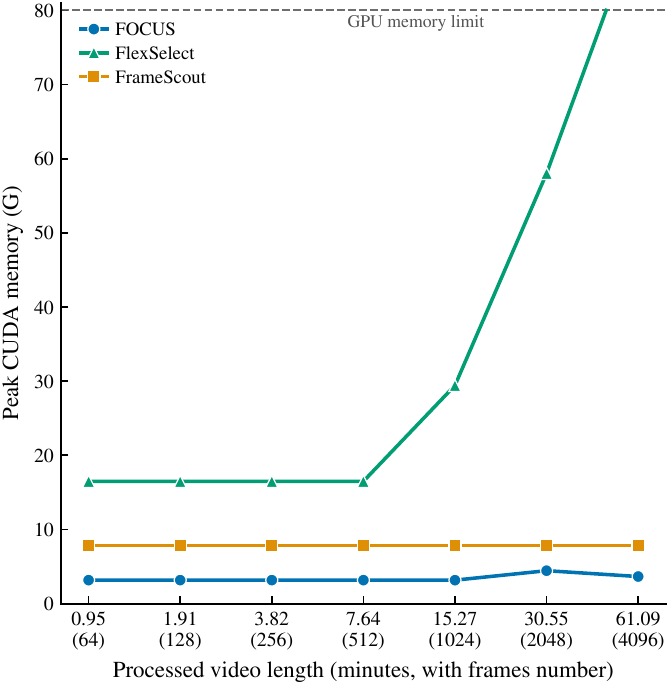}
        \captionof{figure}{Memory vs.\ video length.}
        \label{fig:memory_vs_video_length}
    \end{minipage}
\end{figure}

\textbf{Inference Efficiency and Scalability.}
In practice, users often pose multiple queries about the same video
(\eg, first requesting a summary, then drilling into specific scenes),
making multi-turn QA a common and important use case.
We therefore benchmark the end-to-end inference latency of 
\methodName{}
against FOCUS~\citep{zhu2025focus} and FlexSelect~\citep{zhang2025flexselect}
as the number of queries on the same video grows from $1$ to $8$
(Figure~\ref{fig:efficiency_comparison}).
The efficiency gap widens sharply as queries accumulate.
Both FOCUS 
and FlexSelect must re-encode the video for every new query,
so their total latency grows linearly with the number of queries.
In contrast, \methodName{} computes frame embeddings once and reuses them
across all queries, keeping total latency nearly constant.
At eight queries, \methodName{} achieves a $21.2\times$ speedup over FOCUS
and a $6.3\times$ speedup over FlexSelect.

We further measure peak GPU memory of frame selection process as video length increases from 64 to 4096 frames (Figure~\ref{fig:memory_vs_video_length}).
FlexSelect's memory grows and exceeds $80$\,GB at $4096$ frames, making it impractical for hour-scale videos on a single GPU.
In contrast, \methodName{} maintains constant memory (${\sim}8$\,GB) regardless of video length, consistent with the $\hO((L_s + WBL)\,d)$ bound derived in Section~\ref{sec:memory}, and remains $10\times$ lower than FlexSelect at $4096$ frames.
\rev{Moreover, transferring the retained frame embeddings between GPU and disk for reuse takes only 35.75\,ms for an 8192-frame video, merely $0.014\%$ of its 252\,s encoding time, which is negligible.}

\subsection{Analysis}
\label{subsec:ablation}

\subsubsection{Effectiveness of the Streaming Selector Architecture}
\label{sec:ablation_kv_cache}

We validate the streaming selector architecture
(Section~\ref{sec:memory}) by comparing against a
\textit{Separate Windows} baseline that partitions the video into
non-overlapping chunks of $B$ frames and processes each chunk
independently, with no temporal context propagated across chunk
boundaries. All other components remain identical.

As shown in Table~\ref{tab:kv_cache_ablation}, the streaming
architecture consistently improves accuracy across both VideoLLMs and
benchmarks.
This confirms that the streaming architecture produces frame
representations with richer cross-boundary temporal context, which is
critical for queries requiring reasoning over events that span
multiple chunks.
\rev{Finer-grained ablations on successor aggregation and the contrastive objective, together with analyses on frame budgets, pseudo-label reliability, and other design choices, are provided in Appendix~\ref{apd:component_ablation}--\ref{apd:bolt_prompt}.}

\begin{figure}[!t]
    \centering
    \begin{minipage}[c]{0.5\textwidth}
        \centering
        \captionof{table}{Ablation study on the streaming selector architecture.}
        \label{tab:kv_cache_ablation}
        \resizebox{\linewidth}{!}{
        \begin{NiceTabular}{cccc}
            \CodeBefore
            \rectanglecolor{Gray}{3-2}{3-4}
            \rectanglecolor{Gray}{5-2}{5-4}
            \Body
            \toprule
            \textbf{VideoLLM $\hM$} & \textbf{Method} & \textbf{VideoMME} & \textbf{LongVB} \\
            \midrule
            \multirow{2}{*}{Qwen2.5-VL-7B}
            & Separate Windows & 65.0 & 61.3 \\
            & Streaming   & \textbf{65.6} & \textbf{63.3} \\
            \midrule
            \multirow{2}{*}{LLaVA-Video-7B}
            & Separate Windows & 63.7 & 60.6 \\
            & Streaming   & \textbf{65.7} & \textbf{63.4} \\
            \bottomrule
        \end{NiceTabular}
        }
    \end{minipage}
    \hfill
    \begin{minipage}[c]{0.47\textwidth}
        \centering
        \captionof{table}{Generalization across selector backbones on VideoMME using Qwen2.5-VL-7B.}
        \label{tab:selector_ablation}
        \resizebox{\linewidth}{!}{
        \begin{tabular}{lcc}
            \toprule
            \textbf{Selector $\hS_\vtheta$} & \textbf{Long} & \textbf{Overall} \\
            \midrule
            Uniform (no selector) & 51.4 & 62.6 \\ 		\rowcolor{Gray}
            \methodName{} (LLaVA-OV-0.5B) & 55.8 & 64.6 \\ 		\rowcolor{Gray}
            \methodName{} (InternVL2.5-1B) & \textbf{57.3} & \textbf{65.6} \\
            \bottomrule
        \end{tabular}
        }
    \end{minipage}
\end{figure}

\subsubsection{Generalization Across Selector Backbones}
\label{sec:ablation_backbone}

The selector is designed to be architecture-agnostic:
any lightweight VideoLLM can serve as its backbone.
To verify this, we replace the default InternVL2.5-1B backbone with
LLaVA-OneVision-0.5B~\citep{li2024llavaonevisioneasyvisualtask},
a model with a different architecture and less than half the parameters.
As shown in Table~\ref{tab:selector_ablation}, both backbones yield
substantial improvements over uniform sampling, confirming that the
gains stem from the streaming architecture and contrastive training
framework (Sections~\ref{sec:memory}--\ref{sec:training}) rather than
from a specific backbone choice.
The larger InternVL2.5-1B selector provides additional gains, suggesting
that increased selector capacity further improves frame selection quality.

\subsubsection{Representational Consistency}
\label{sec:consistency}

\begin{wrapfigure}{r}{0.35\textwidth}
    \vskip -.3in
    \centering
    \includegraphics[width=0.88\linewidth]{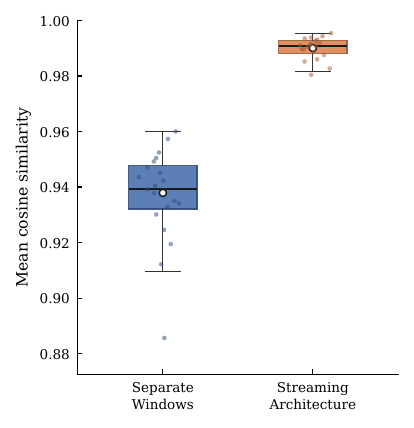}
    \vskip -.1in
    \caption{Distribution of similarities between two strategies.}
    \label{fig:consistency}
\end{wrapfigure}

The streaming architecture processes chunks sequentially under a bounded KV cache,
raising a natural question: do the resulting frame representations degrade
compared to full-sequence inference that processes all frames at once?
To answer this, we compare the two on $100$ videos each containing $N=1024$ frames,
computing the mean frame-level cosine similarity between their outputs per video
and visualizing the distribution in Figure~\ref{fig:consistency}.

As shown, the streaming architecture yields per-video similarities
tightly concentrated near $1.0$, closely recovering the full-sequence
representations under constant KV-cache memory. 
In contrast, Separate
Windows exhibits substantially lower similarity with a larger variance, confirming that processing chunks in isolation leads to
significant representational degradation.

\begin{revblock}
\subsubsection{Training Context Strategy}
\label{sec:train_context}

As described in Section~\ref{sec:training_detail}, the selector is trained on independent chunks with only the sink KV cache, while the sliding-window KV cache is activated only at inference.
To test whether this training-inference difference hurts performance, we train two selectors from the same initialization under the same wall-clock budget, changing only the training context: independent chunks (ours) versus the streaming KV context used at inference.
As shown in Table~\ref{tab:train_context}, independent-chunk training converges faster (lower training loss) and achieves higher accuracy on both benchmarks, confirming that our training strategy is both more efficient and more accurate.
\end{revblock}

\begin{table}[!h]
    \centering
    \begin{revblock}
    \caption{\rev{Comparison of training context strategies under the same wall-clock budget using Qwen2.5-VL-7B.}}
    \label{tab:train_context}
    \begin{tabular}{lccc}
        \toprule
        \textbf{Training context} & \textbf{Training loss} & \textbf{VideoMME} & \textbf{LongVB} \\
        \midrule
        Streaming KV context & 0.3731 & 64.2 & 62.1 \\
        \rowcolor{Gray} Independent chunks (ours) & \textbf{0.2057} & \textbf{65.6} & \textbf{63.3} \\
        \bottomrule
    \end{tabular}
    \end{revblock}
\end{table}

\subsubsection{Visualization of the Shared Embedding Space}
\label{sec:tsne_visualization}

To verify that the frame-query contrastive objective
(Section~\ref{sec:training}) learns a well-structured shared embedding
space, we use t-SNE~\citep{van2008visualizing} to project the frame and query embeddings into a 2D
plane. We select a sample from the
NExT-GQA dataset~\citep{nextgqa} with ground-truth (GT) temporal
annotations.

\begin{figure}[!h]
    \centering
    \includegraphics[width=\linewidth]{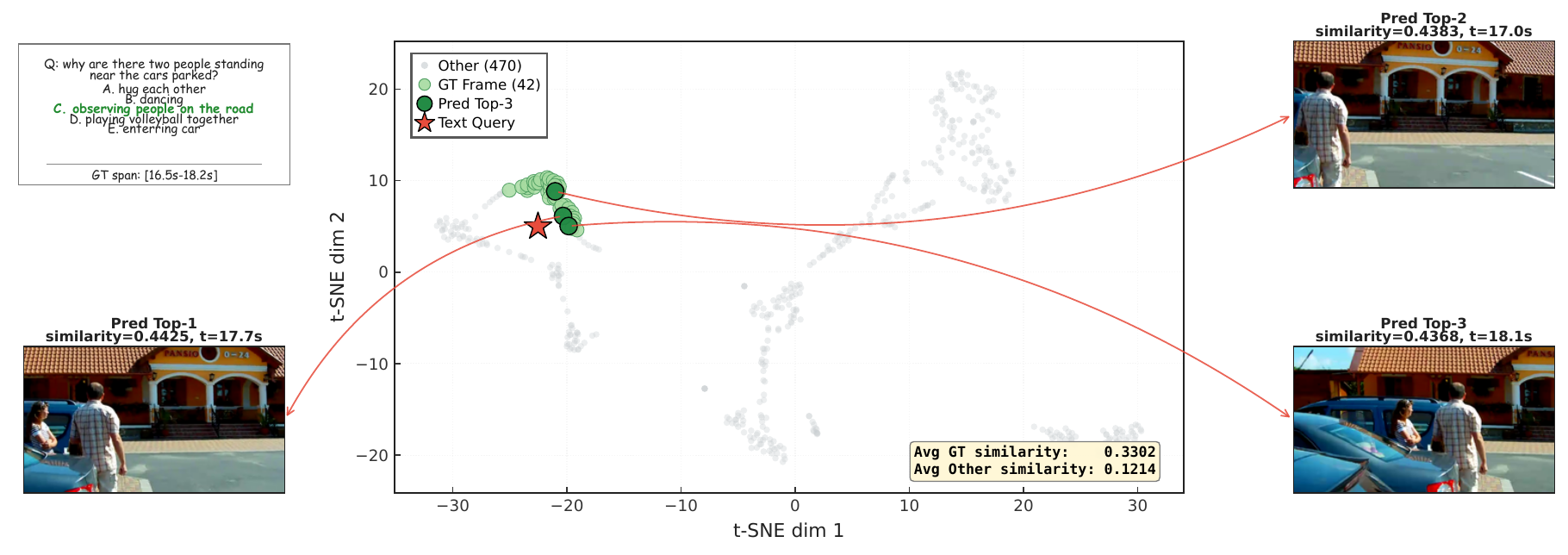}
    \caption{
        t-SNE visualization of query and frame embeddings.
    }
    \label{fig:tsne_embedding}
\end{figure}

As can be seen from Figure~\ref{fig:tsne_embedding},
the query embedding (red star) is mapped closer to the GT keyframes
than to the background frames, with an average cosine similarity of
$0.33$ for keyframes versus $0.12$ for background frames. Moreover, the
top-3 retrieved frames fall precisely within the GT span, demonstrating
that the contrastive objective successfully separates query-relevant
frames from irrelevant ones in the shared space, enabling accurate
retrieval via simple cosine similarity.

\subsubsection{Qualitative Results}
\label{sec:qualitative_results}

\begin{figure}[!h]
    \centering
    \includegraphics[width=0.85\linewidth]{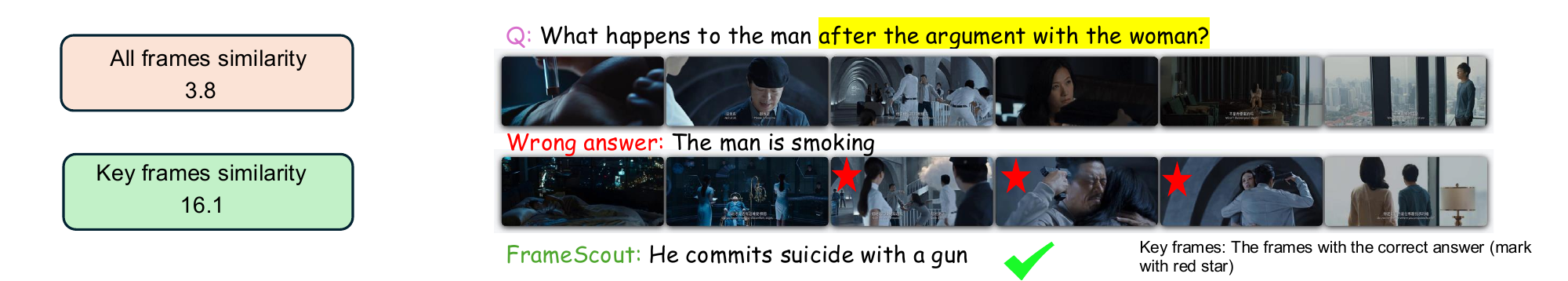}
    \caption{
        Qualitative comparison with FlexSelect on a temporal reasoning query.
    }
    \label{fig:qualitative_case}
\end{figure}

Figure~\ref{fig:qualitative_case} shows a qualitative comparison with
FlexSelect~\citep{zhang2025flexselect} on a temporal reasoning query. 
As can be seen,
FlexSelect fails to propagate the anchoring event
across chunk boundaries and selects temporally irrelevant frames,
leading to the wrong answer. 
In contrast, \methodName{} successfully selects the key post-event frames marked with red stars
and generates the correct answer, verifying that the streaming architecture captures cross-chunk temporal dependencies that isolated per-chunk encoding misses. The higher cosine similarity of the selected keyframes further confirms that the contrastive objective learns a well-separated embedding space where query-relevant frames are clearly distinguished from irrelevant ones. Additional examples are provided in Appendix~\ref{sec:additional examples}.

\section{Conclusion}

In this paper, we proposed \methodName, a keyframe selection framework for long video
understanding with two core designs: a streaming selector architecture
that produces temporally-contextualized frame embeddings under constant
KV-cache memory and linear computation, and a frame-query contrastive
objective that aligns frame and query embeddings in a shared space for
accurate selection. Experiments on three long-video benchmarks and six
base VideoLLMs demonstrate consistent state-of-the-art results, while achieving a
$6\times$ speedup in multi-query scenarios.

\section*{Acknowledgement}

This research was supported in part by National Key Research and Development Program of China under Grant no. 2022YFA1004102, National Natural Science Foundation of China under Grant no. 62136005, National Science Fund for Distinguished Young Scholars no. 62325204, Shenzhen fundamental research program JCYJ20250604144724032, and the Research Grants Council of the Hong Kong Special Administrative Region (Grants 16202523 and 16213126).

{
\small
\bibliography{refs}
\bibliographystyle{abbrvnat}
}

\newpage
\appendix

\section{Training and Inference Algorithms}
\label{apd:algorithms}

We provide the complete training and inference procedures of \methodName in Algorithm~\ref{alg:train} and Algorithm~\ref{alg:infer}, respectively.


\begin{algorithm}[!h]
    \caption{\methodName: Training via frame-query contrastive learning.}
    \label{alg:train}
    \begin{algorithmic}[1]
    \Require Dataset $\mathcal{D}=\{(\vF,Q, A)\}$; teacher VideoLLM $\mathcal{M}_{\text{teacher}}$;
             trainable $\vtheta=\{\mathcal{G},\phi\}$ ($\mathcal{E}$ frozen);
             training frame count $N'$;
             chunk size $B$; pseudo-label top-$k$;
             contrastive temperature $\tau$;
             successor discount $\beta$; learning rate $\eta$;
             sink KV cache $\text{KV}(\mathbf{s})$.
    \Repeat
      \State Sample a mini-batch $\hB$ from $\mathcal{D}$;
      \For{each $(\vF,Q, A)\in \hB$}
          \Statex \hspace{1.5em}\textbf{Pseudo-label generation:}
          \State $\vF' = (F_1,\ldots,F_{N'}) \gets$ uniformly subsample $N'$ frames from $\vF$;
          \State $(\mathcal{T}_+,\mathcal{T}_-) \gets \textsc{TeacherPseudoLabel}(\mathcal{M}_{\text{teacher}},\vF',Q,k)$;
          \Statex \hspace{1.5em}\textbf{Forward encoding:}
          \For{$i=1,\ldots,N'$}
              \State $\mathbf{v}_i \gets \mathcal{E}(F_i) \in \mathbb{R}^{L \times d}$;
          \EndFor
          \For{$t=1,\ldots,N'/B$}
              \State $\mathbf{V}_t \gets (\mathbf{v}_{(t-1)B+1},\ldots,\mathbf{v}_{tB})$;
              \State $\big(\vh_{(t-1)B+1},\ldots,\vh_{tB}\big) \gets \mathcal{G}\!\left(\text{KV}(\mathbf{s}),\,\mathbf{V}_t\right)$;
          \EndFor
          \Statex \hspace{1.5em}\textbf{Successor aggregation:}
          \State $\mathbf{z}_{N'} \gets \vh_{N'}$;
          \For{$i=N'-1,\ldots,1$}
              \State $\mathbf{z}_i \gets \vh_i + \beta\,\mathbf{z}_{i+1}$;
          \EndFor
          \State $\vM_i \gets \phi(\mathbf{z}_i)$ for all $i=1,\ldots,N'$;
          \Statex \hspace{1.5em}\textbf{Query encoding and contrastive loss:}
          \State $\vM_Q \gets \phi(\mathcal{G}(Q))$;
          \State Compute $\mathcal{L}$ via Eq.~\eqref{eq:infonce};
      \EndFor
      \State $\vtheta \gets \vtheta - \eta\,\nabla_{\vtheta}\mathcal{L}$;
    \Until{convergence}
    \State \Return $\mathcal{S}_\vtheta=\{\mathcal{E},\mathcal{G},\phi\}$.
    \end{algorithmic}
\end{algorithm}

\begin{algorithm}[!h]
    \caption{\methodName: Inference via streaming encoding and per-query retrieval.}
    \label{alg:infer}
    \begin{algorithmic}[1]
    \Require Video $\vF=(F_1,\ldots,F_N)$ ($N$ a multiple of $B$); queries $\{Q_j\}_{j=1}^m$;
             trained selector $\mathcal{S}_\vtheta=\{\mathcal{E},\mathcal{G},\phi\}$;
             base VideoLLM $\mathcal{M}$;
             chunk size $B$; sliding-window size $W$;
             successor discount $\beta$; frame budget $K$;
             sink KV cache $\text{KV}(\mathbf{s})$.
    \Statex \textbf{Stage 1: Streaming video encoding (once per video):}
    \For{$i=1,\ldots,N$}
        \State $\mathbf{v}_i \gets \mathcal{E}(F_i) \in \mathbb{R}^{L \times d}$;
    \EndFor
    \State Initialize sliding buffer $\text{KV}_{\text{win}} \gets \emptyset$;
    \For{$t = 1, \ldots, N/B$}
        \State $\mathbf{V}_t \gets (\mathbf{v}_{(t-1)B+1},\ldots,\mathbf{v}_{tB})$;
        \State $\big(\vh_{(t-1)B+1}, \ldots, \vh_{tB}\big)
                \gets \mathcal{G}\!\left(\text{KV}(\mathbf{s}),\, \text{KV}_{\text{win}},\, \mathbf{V}_t\right)$;
        \State Append $\text{KV}(\mathbf{V}_t)$ to $\text{KV}_{\text{win}}$;
               evict the oldest chunk if $|\text{KV}_{\text{win}}| > W$;
    \EndFor
    \Statex \hspace{1.5em}\textbf{Successor aggregation:}
    \State $\mathbf{z}_{N} \gets \vh_{N}$;
    \For{$i=N-1,\ldots,1$}
        \State $\mathbf{z}_i \gets \vh_i + \beta\,\mathbf{z}_{i+1}$;
    \EndFor
    \State $\vM_i \gets \phi(\mathbf{z}_i)$ for all $i=1,\ldots,N$;
    
    \Statex \textbf{Stage 2: Per-query retrieval and answering:}
    \For{$j = 1, \ldots, m$}
        \State $\vM_{Q_j} \gets \phi(\mathcal{G}(Q_j))$;
        \State $\hat{\mathcal{I}}_j \gets \operatorname{Top\text{-}K}\!\big(\{\operatorname{sim}(\vM_{Q_j},\,\vM_i)\}_{i=1}^N\big)$;
        \State $\hat{\vF}_j \gets (F_i)_{i \in \hat{\mathcal{I}}_j}$;
        \State $A_j \gets \mathcal{M}(\hat{\vF}_j,\, Q_j)$;
    \EndFor
    \State \Return $\{A_j\}_{j=1}^m$.
    \end{algorithmic}
\end{algorithm}

\section{Additional Experiments}

\subsection{Sensitivity to Hyperparameters}
\label{subsubsec:sensitivity}

We examine two key hyperparameters: the chunk size $B$ that controls
how many frames the selector processes per step, and the discount
factor $\beta$ in successor aggregation (Section~\ref{sec:memory}).

\textbf{Chunk size $B$.}
As shown in Table~\ref{tab:chunk_ablation}, varying $B$ from 8 to 16
on Qwen3-VL-8B causes VideoMME overall to fluctuate by only 0.5,
indicating that \methodName{} is robust to this choice.
We use $B{=}12$ as the default in all experiments.

\textbf{Discount factor $\beta$.}
Table~\ref{tab:beta_ablation} shows the effect of $\beta$ on
LLaVA-Video-7B.
Setting $\beta{=}0$ disables successor aggregation entirely, reducing
frame representations to purely causal (past-only) context;
performance drops by 0.4 on VideoMME Overall compared to $\beta{=}0.3$,
confirming that incorporating future context is beneficial.
However, a large $\beta$ (\eg, 0.8) over-weights distant future frames
and degrades accuracy.
$\beta{=}0.3$ is used in all experiments as it balances the future and
past context.

\begin{table}[!h]
    \centering
    \caption{
        Sensitivity to chunk size $B$
        on VideoMME using Qwen3-VL-8B.
    }
    \label{tab:chunk_ablation}
    \setlength{\tabcolsep}{8pt}
    \renewcommand{\arraystretch}{1.2}
    \begin{tabular}{ccc}
        \toprule
        \textbf{Chunk size $B$} & \textbf{Long} & \textbf{Overall} \\
        \midrule
        8  & 60.3 & 68.8 \\ 		\rowcolor{Gray}
        12 & \textbf{61.2} & \textbf{69.3} \\
        16 & 61.1 & 68.9 \\
        \bottomrule
    \end{tabular}
\end{table}

\begin{table}[!h]
    \centering
    \caption{
        Sensitivity to discount factor $\beta$ on VideoMME using LLaVA-Video-7B.
    }
    \label{tab:beta_ablation}
    \renewcommand{\arraystretch}{1.15}
    \begin{tabular}{ccc}
        \toprule
        $\beta$ & \textbf{Long} & \textbf{Overall} \\
        \midrule
        0.0 & 55.5 & 65.3 \\
        0.2 & 55.1 & 65.5 \\ 		\rowcolor{Gray}
        0.3 & \textbf{56.1} & \textbf{65.7} \\
        0.5 & 54.6 & 64.8 \\
        0.8 & 51.4 & 63.2 \\
        \bottomrule
    \end{tabular}
\end{table}

\begin{revblock}

\subsection{Comparison with More Baseline Methods}
\label{apd:recent_methods}

We further compare \methodName{} with the recent training-free selectors BOLT~\citep{liu2025bolt} and F2C~\citep{sun2025f2c}, the training-based selector M-LLM~\citep{hu2025mllm}, and the token compression methods HICom~\citep{liu2025hicom} and VTM~\citep{lee2024vtm}, under the same protocol as Table~\ref{tab:main_results}; M-LLM is compared under its Qwen2-VL-7B~\citep{wang2024qwen2vl} setting.
As shown in Table~\ref{tab:recent_methods}, \methodName{} outperforms the training-free selectors BOLT and F2C (\eg, $+2.1$ on VideoMME and $+1.7$ on LongVB over BOLT), the training-based selector M-LLM ($+4.1$ on VideoMME and $+3.6$ on LongVB), and the token compression methods HICom and VTM ($+7.8$ and $+4.5$ on VideoMME).

\begin{table}[!h]
    \centering
    \begin{revblock}
    \caption{\rev{Comparison with recent training-free selectors (F2C, BOLT), token compression methods (HICom, VTM), and the training-based selector M-LLM. VideoMME reports Overall accuracy. ``--'': not evaluated.}}
    \label{tab:recent_methods}
    \begin{tabular}{llcc}
        \toprule
        $\hM$ & \textbf{Method} & \textbf{VideoMME} & \textbf{LongVB} \\
        \midrule
        \multirow{5}{*}{\rotatebox[origin=c]{90}{\scriptsize Qwen2.5-VL}}
        & F2C~\citep{sun2025f2c} & 62.3 & 60.6 \\
        & BOLT~\citep{liu2025bolt} & 63.5 & 61.6 \\
        & HICom~\citep{liu2025hicom} & 57.8 & -- \\
        & VTM~\citep{lee2024vtm} & 61.1 & -- \\
        & \cellcolor{Gray}\methodName & \cellcolor{Gray}\textbf{65.6} & \cellcolor{Gray}\textbf{63.3} \\
        \midrule
        \multirow{2}{*}{\rotatebox[origin=c]{90}{\scriptsize Qwen2}}
        & M-LLM~\citep{hu2025mllm} & 58.7 & 57.0 \\
        & \cellcolor{Gray}\methodName & \cellcolor{Gray}\textbf{62.8} & \cellcolor{Gray}\textbf{60.6} \\
        \bottomrule
    \end{tabular}
    \end{revblock}
\end{table}

\subsection{Contribution of Each Component}
\label{apd:component_ablation}

\methodName{} consists of three components: the sliding-window KV cache and successor aggregation in the streaming selector architecture (Section~\ref{sec:memory}), and the frame-query contrastive objective (Section~\ref{sec:training}).
To disentangle their contributions, we compare \methodName{} with three variants, each changing exactly one component: removing the sliding-window KV cache (\ie, the Separate Windows baseline in Section~\ref{sec:ablation_kv_cache}), removing successor aggregation, or replacing the contrastive objective with the Spearman rank-correlation loss used in FlexSelect~\citep{zhang2025flexselect}.
As shown in Table~\ref{tab:component_ablation}, the contrastive objective is the dominant contributor: replacing it with the Spearman loss causes the largest drop ($-4.5$ on VideoMME and $-2.6$ on LongVB). Removing either architectural component yields smaller but consistent degradation on both benchmarks.

\begin{table}[!h]
    \centering
    \caption{Effect of components in \methodName{} using Qwen2.5-VL-7B.}
    \label{tab:component_ablation}
    \begin{tabular}{lcc}
        \toprule
        \textbf{Variant} & \textbf{VideoMME} & \textbf{LongVB} \\
        \midrule
        w/o sliding-window KV cache (Separate Windows) & 65.0 & 61.3 \\
        w/o successor aggregation & 65.3 & 62.7 \\
        w/o contrastive objective (Spearman) & 61.1 & 60.7 \\
        \rowcolor{Gray} Full \methodName & \textbf{65.6} & \textbf{63.3} \\
        \bottomrule
    \end{tabular}
\end{table}

\subsection{Robustness to Frame Budget}
\label{apd:frame_budget}

\textbf{Low budgets.}
We evaluate all selectors with $K\in\{4,8,16,60\}$ (Table~\ref{tab:frame_budget}).
\methodName{} achieves the best accuracy in every setting, and its advantage is most pronounced under low budgets ($+5.9$ over Uniform at $K{=}4$ versus $+3.0$ at $K{=}60$ on VideoMME Overall).
A small $K$ demands high precision at the top of the ranking, which our contrastive objective directly optimizes by separating relevant from irrelevant frames, matching the top-$K$ criterion at inference.

\begin{table}[!h]
    \centering
    \begin{revblock}
    \caption{\rev{Accuracy under different frame budgets $K$ on VideoMME using Qwen2.5-VL-7B.}}
    \label{tab:frame_budget}
    \resizebox{0.8\linewidth}{!}{
    \begin{tabular}{lcccccccc}
        \toprule
        & \multicolumn{4}{c}{\textbf{Overall}} & \multicolumn{4}{c}{\textbf{Long}} \\
        \cmidrule(lr){2-5} \cmidrule(lr){6-9}
        \textbf{Method} & $K{=}4$ & $8$ & $16$ & $60$ & $K{=}4$ & $8$ & $16$ & $60$ \\
        \midrule
        Uniform    & 49.2 & 53.9 & 57.3 & 62.6 & 44.4 & 47.8 & 48.3 & 51.4 \\
        FOCUS      & 54.4 & 57.0 & 57.7 & 63.4 & 47.4 & 47.3 & 49.0 & 54.3 \\
        FlexSelect & 52.7 & 55.0 & 59.0 & 64.9 & 47.6 & 48.3 & 51.8 & 55.4 \\
        \rowcolor{Gray} \methodName & \textbf{55.1} & \textbf{57.4} & \textbf{60.3} & \textbf{65.6} & \textbf{48.1} & \textbf{51.4} & \textbf{52.7} & \textbf{57.3} \\
        \bottomrule
    \end{tabular}
    }
    \end{revblock}
\end{table}

\textbf{High budget (1\,fps).}
We further evaluate a high-budget regime at 1\,fps on VideoMME videos shorter than 10 minutes, so that all selected frames fit in the context capacity of the base VideoLLM.
All methods rank the same candidate frames, with $K$ set by the video duration to yield one selected frame per second.
As shown in Table~\ref{tab:fps1}, \methodName{} still achieves the best accuracy among all methods.

\begin{table}[!h]
    \centering
    \caption{\rev{Accuracy (\%) under a 1\,fps frame budget on the VideoMME subset with videos shorter than 10 minutes, using Qwen2.5-VL-7B.}}
    \label{tab:fps1}
    \begin{tabular}{lcccc}
        \toprule
        & \textbf{Uniform} & \textbf{FOCUS} & \textbf{FlexSelect} & \textbf{\methodName} \\
        \midrule
        Accuracy & 64.9 & 61.4 & 65.3 & \cellcolor{Gray}\textbf{66.2} \\
        \bottomrule
    \end{tabular}
\end{table}

\subsection{Reliability of Teacher-Attention Pseudo-Labels}
\label{apd:pseudo_label}

To validate the supervision signal, we measure how often the teacher-attention pseudo-labels hit the human-annotated ground-truth (GT) temporal spans of NExT-GQA~\citep{nextgqa}.
As shown in Table~\ref{tab:pseudo_label}, more than $99\%$ of GT spans are recalled by the top-40 pseudo-labeled frames, confirming that teacher-attention scores reliably correspond to answer-relevant frames identified by human annotators.
We further show in Appendix~\ref{apd:teacher} that \methodName{} is insensitive to the choice of teacher.

\begin{table}[!h]
    \centering
    \caption{\rev{Recall (\%) of human-annotated GT temporal spans on NExT-GQA by the top-$K$ teacher-attention pseudo-labeled frames.}}
    \label{tab:pseudo_label}
    \begin{tabular}{lcccc}
        \toprule
        Top-$K$ & 10 & 20 & 30 & 40 \\
        \midrule
        GT span recall & 73.6 & 86.0 & 91.7 & \textbf{99.2} \\
        \bottomrule
    \end{tabular}
\end{table}

\subsection{Design of the Projection Head}
\label{apd:projector}

We compare three projection heads $\phi$: \emph{Identity} (directly using the backbone hidden states), \emph{Linear} (a single trainable linear layer), and our default \emph{two-layer MLP}.
As shown in Table~\ref{tab:projector}, the MLP performs best on all benchmarks, \eg, $+3.9$ on VideoMME Overall over Identity, showing that hidden states pretrained for next-token prediction are not directly suited for similarity-based retrieval. The MLP also consistently outperforms Linear, indicating that a nonlinear projection is beneficial for reshaping frame and query representations into a shared retrieval space.

\begin{table}[!h]
    \centering
    \caption{\rev{Ablation on the projection head $\phi$ using Qwen2.5-VL-7B.}}
    \label{tab:projector}
    \begin{tabular}{lccc}
        \toprule
        \textbf{Projector} & \textbf{VideoMME Overall} & \textbf{VideoMME Long} & \textbf{LongVB} \\
        \midrule
        Identity & 61.7 & 56.4 & 60.3 \\
        Linear & 63.0 & 56.0 & 59.9 \\
        \rowcolor{Gray} Two-layer MLP (ours) & \textbf{65.6} & \textbf{57.3} & \textbf{63.3} \\
        \bottomrule
    \end{tabular}
\end{table}

\subsection{Choice of Future Context Aggregation}
\label{apd:future_agg}

We compare successor aggregation with two simpler future-context alternatives: \emph{backward temporal smoothing}, which averages the hidden states over a fixed-size future window, and \emph{non-causal temporal aggregation}, which uniformly averages all future hidden states. Note that exponential future pooling is mathematically identical to successor aggregation (Eq.~3) and is thus not a separate baseline.
As shown in Table~\ref{tab:future_agg}, successor aggregation outperforms both alternatives on both benchmarks, confirming that the exponential decay structure, which weights nearby frames more than distant ones, is essential.

\begin{table}[!h]
    \centering
    \caption{\rev{Comparison of future context aggregation strategies using Qwen2.5-VL-7B.}}
    \label{tab:future_agg}
    \begin{tabular}{lcc}
        \toprule
        \textbf{Future aggregation} & \textbf{VideoMME Overall} & \textbf{LongVB} \\
        \midrule
        Backward temporal smoothing & 65.0 & 61.3 \\
        Non-causal temporal aggregation & 65.2 & 62.4 \\
        \rowcolor{Gray} Successor aggregation (ours) & \textbf{65.6} & \textbf{63.3} \\
        \bottomrule
    \end{tabular}
\end{table}

\subsection{Sensitivity to the Teacher Model}
\label{apd:teacher}

We re-generate the pseudo-labels with a different teacher, Qwen2.5-VL-7B, and retrain the selector under the same setting.
As shown in Table~\ref{tab:teacher}, the two teachers yield close downstream accuracy (average $\Delta = 0.3$), and both outperform FlexSelect, showing that \methodName{} is insensitive to the choice of teacher.

\begin{table}[!h]
    \centering
    \caption{\rev{Sensitivity to the teacher model used for pseudo-label generation using Qwen2.5-VL-7B.}}
    \label{tab:teacher}
    \setlength{\tabcolsep}{4pt}
    \begin{tabular}{lccc}
        \toprule
        \textbf{Method} & \textbf{VideoMME Long} & \textbf{VideoMME Overall} & \textbf{LongVB} \\
        \midrule
        FlexSelect~\citep{zhang2025flexselect} & 55.4 & 64.9 & 63.0 \\
        \rowcolor{Gray} \methodName{} (teacher: InternVL2.5-8B) & \textbf{57.3} & \textbf{65.6} & \textbf{63.3} \\
        \methodName{} (teacher: Qwen2.5-VL-7B) & 57.0 & 65.2 & 63.1 \\
        \bottomrule
    \end{tabular}
\end{table}

\subsection{Adaptive Frame Budget}
\label{apd:adaptive_k}

Instead of selecting a fixed number of frames, we select every frame whose cosine similarity to the query embedding exceeds a threshold $\delta = 0.7$, without any retraining.
As shown in Table~\ref{tab:adaptive_k}, the adaptive variant achieves higher accuracy on LongVB while using fewer frames on average, suggesting that the number of relevant frames varies across queries and that adapting $K$ is a promising direction.

\begin{table}[!h]
    \centering
    \caption{\rev{Fixed versus threshold-based adaptive frame selection on LongVB using Qwen2.5-VL-7B.}}
    \label{tab:adaptive_k}
    \begin{tabular}{lcc}
        \toprule
        \textbf{Selection rule} & \textbf{Accuracy} & \textbf{Avg. \#Frames} \\
        \midrule
        Fixed $K$ & 61.7 & 30 \\
        \rowcolor{Gray} Adaptive ($\operatorname{sim} > \delta$) & \textbf{62.0} & \textbf{26.5} \\
        \bottomrule
    \end{tabular}
\end{table}

\subsection{Robustness to Evaluation Prompts}
\label{apd:bolt_prompt}

For a fair comparison, all methods in Table~\ref{tab:main_results} and Table~\ref{tab:recent_methods} are evaluated with the same prompt template.
To verify that the gains of \methodName{} do not depend on this template, we further compare \methodName{} with BOLT~\citep{liu2025bolt} under the prompt from BOLT's official repository, keeping all other components (frame sampling, frame selection, and answer extraction) unchanged.
As shown in Table~\ref{tab:bolt_prompt}, \methodName{} outperforms BOLT under both prompts and all frame budgets, \eg, by $+2.07$ under our prompt and $+2.11$ under BOLT's official prompt at $K{=}60$, showing that its advantage is robust to the evaluation prompt.
Under BOLT's official prompt, our reproduced BOLT achieves 64.19 at $K{=}32$, consistent with its officially reported 64.0.

\begin{table}[!h]
    \centering
    \caption{\rev{BOLT and \methodName{} under our evaluation prompt and BOLT's official prompt on VideoMME Overall using Qwen2.5-VL-7B.}}
    \label{tab:bolt_prompt}
    \begin{tabular}{llcccc}
        \toprule
        \textbf{Prompt} & \textbf{Method} & $K{=}8$ & $K{=}16$ & $K{=}32$ & $K{=}60$ \\
        \midrule
        \multirow{2}{*}{Ours} & BOLT~\citep{liu2025bolt} & 55.63 & 59.44 & 62.07 & 63.52 \\
        & \cellcolor{Gray}\methodName & \cellcolor{Gray}\textbf{57.44} & \cellcolor{Gray}\textbf{60.30} & \cellcolor{Gray}\textbf{63.11} & \cellcolor{Gray}\textbf{65.59} \\
        \midrule
        \multirow{2}{*}{BOLT official} & BOLT~\citep{liu2025bolt} & 57.44 & 61.07 & 64.19 & 64.67 \\
        & \cellcolor{Gray}\methodName & \cellcolor{Gray}\textbf{59.07} & \cellcolor{Gray}\textbf{61.26} & \cellcolor{Gray}\textbf{64.67} & \cellcolor{Gray}\textbf{66.78} \\
        \bottomrule
    \end{tabular}
\end{table}
\end{revblock}

\subsection{Additional Examples}
\label{sec:additional examples}

\begin{figure}[!h]
    \centering
    \includegraphics[width=0.95\linewidth]{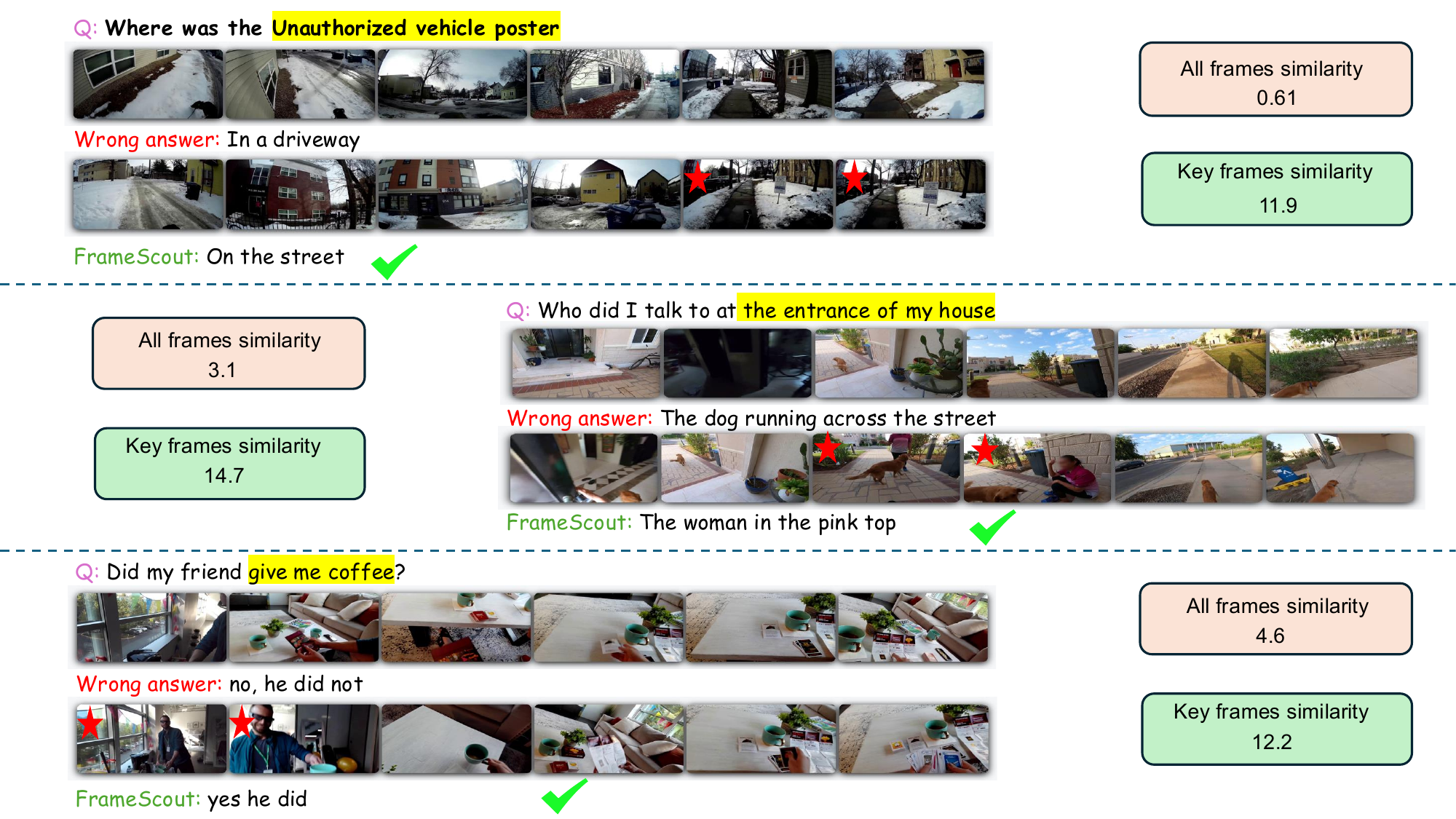}
    \caption{
        Comparison with the frame selection results of the FOCUS~\citep{zhu2025focus}.
    }
    \label{fig:qualitative_case-x1}
\end{figure}

\begin{figure}[!h]
    \centering
    \includegraphics[width=0.95\linewidth]{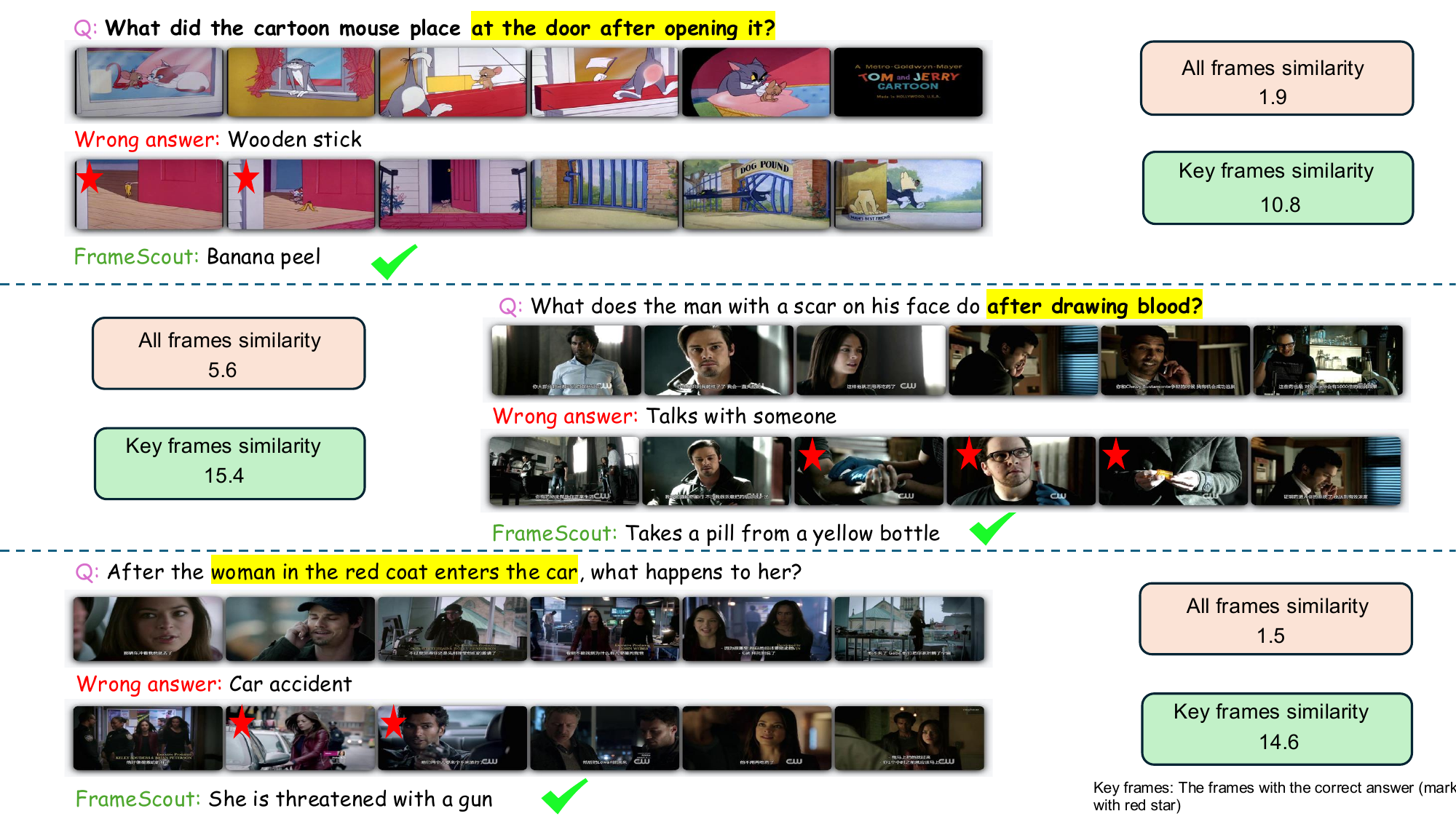}
    \caption{
        Comparison with the frame selection results of the FlexSelect~\citep{zhang2025flexselect}.
    }
    \label{fig:qualitative_case-x2}
\end{figure}

We provide additional qualitative comparisons with FOCUS~\citep{zhu2025focus} (Figure~\ref{fig:qualitative_case-x1}) and FlexSelect~\citep{zhang2025flexselect} (Figure~\ref{fig:qualitative_case-x2}).
In both cases, the baselines select temporally irrelevant frames due to the lack of cross-chunk temporal context, leading to incorrect answers.
In contrast, \methodName{} consistently selects the query-relevant frames, confirming that the streaming architecture captures temporal dependencies across chunk boundaries and the contrastive objective produces a well-separated embedding space for accurate frame selection.

\section{Additional Related Works}
\label{sec:additional_related_works}

\textbf{Video Large Language Models.}
Early VideoLLMs~\citep{zhang2024llava,chen2024internvl,bai2025qwen3,qwen2.5-VL} extend
image-based multimodal LLMs by densely encoding a small number of sampled frames,
yielding strong performance on short video benchmarks but scaling poorly with
video duration.
To enable long-video understanding, three main lines of work have emerged.
\emph{Context-window extension}~\citep{xue2024longvila,zhang2024longva} scales the
underlying LLM to hundreds of thousands of tokens, allowing denser frame sampling
at the cost of expensive long-context training and inference.
\emph{Visual token compression}~\citep{li2024llamavid,shu2024videoxl} reduces the
per-frame token count via learned aggregation or summarization tokens, but risks
discarding fine-grained visual details crucial for question answering.
\emph{Memory-augmented architectures}~\citep{he2024malmm,song2024moviechat}
maintain an online memory bank that stores compressed visual features, decoupling
the model's working memory from total video length.
All these methods modify, retrain, or wrap the base VideoLLM.
In contrast, \methodName operates as an external, lightweight keyframe selector
that is fully decoupled from the base VideoLLM and can be combined with any
open- or closed-source VideoLLM without modification.

\begin{revblock}
\textbf{Large Language Models.}
Large language models (LLMs)~\citep{brown2020language,achiam2023gpt4,touvron2023llama,jiang2023mistral}
have demonstrated remarkable capabilities across a wide range of tasks.
\methodName{} builds on them in two roles: a lightweight LLM serves as the backbone of the selector,
and a downstream (Video)LLM produces the final answer.
Much work improves LLMs themselves, \eg, by enhancing reasoning via chain-of-thought
prompting~\citep{wei2022chain,kojima2022large}, self-consistency~\citep{wang2023selfconsistency},
verifiers~\citep{cobbe2021training,jiang2024fobar}, and data bootstrapping~\citep{yu2024metamath},
or by extending them to visual inputs~\citep{liu2023visual,wei2024gita}.
Two lines of work are most related to ours.
\emph{(i) Lightweight modules serving a frozen LLM.}
Parameter-efficient methods adapt a frozen model with soft prompts~\citep{lester2021power,jiang2023metaprompter}
or low-rank adapters~\citep{hu2022lora}, and PARM~\citep{lin2025parm} aligns a frozen LLM at test time
with a small preference-aware reward model, instead of fine-tuning it via RLHF~\citep{ouyang2022training}
or DPO~\citep{rafailov2023direct}.
Similarly, \methodName{} leaves the base VideoLLM untouched, which also avoids the safety degradation
induced by fine-tuning~\citep{qi2024finetuning} that otherwise calls for dedicated
defenses~\citep{jiang2025metadefense,chen2026spard}.
\emph{(ii) Relevance-based selection.}
Mixture-of-experts models route each token to a few experts~\citep{shazeer2017outrageously,fedus2022switch},
MetaMoE~\citep{jiang2026metamoe} selects proxies by relevance and diversity, and
RouterDC~\citep{chen2024routerdc} learns query and LLM embeddings via dual contrastive learning
to route each query to the most suitable LLM.
\methodName{} adopts an analogous query--candidate contrastive alignment, but its candidates are
video frames whose embeddings are computed from content with temporal context, rather than a fixed
set of learnable model embeddings.
While these works improve the LLM itself, \methodName{} improves what it sees, selecting compact
query-relevant evidence from long videos for any (Video)LLM.
\end{revblock}

\begin{revblock}
\section{Limitations}
\label{sec:limitations}

\methodName{} is trained with teacher-attention pseudo-labels, which are only a proxy for true frame-level relevance; higher-quality supervision (\eg, human-annotated evidence frames) could further improve performance.
An adaptive frame budget based on query-specific evidence could further improve efficiency and accuracy, as suggested by our preliminary threshold-based variant (Appendix~\ref{apd:adaptive_k}).
Finally, evicted chunks are retained only implicitly through hidden states propagated along the KV-cache chain; consolidating them into explicit memory tokens kept after the attention sinks is a natural constant-memory extension that we leave to future work.
\end{revblock}

\newpage


\end{document}